\documentclass[10pt,twocolumn,letterpaper]{article}

\usepackage[pagenumbers]{cvpr}
\usepackage{mathpazo}
\usepackage[T1]{fontenc}
\usepackage[scaled=0.95]{inconsolata}
\usepackage{caption}
\usepackage{url}
\usepackage{fancyhdr}
\usepackage{color}
\usepackage[usenames,dvipsnames]{xcolor}
\usepackage{paralist}
\usepackage{mysymbols}
\usepackage{setspace}
\usepackage{graphicx}
\usepackage{longtable}
\usepackage{booktabs}
\usepackage{float}
\usepackage{caption}
\usepackage{subcaption}
\usepackage{dblfloatfix}
\usepackage{amsfonts,amsmath}
\definecolor{bluebell}{RGB}{52,31,151}
\definecolor{amour}{RGB}{238,82,83}
\usepackage[pagebackref=true,breaklinks=true,colorlinks,bookmarks=false,citecolor=bluebell,linkcolor=amour]{hyperref}

\usepackage{epigraph}
\newcommand{\myquote}[1]{\emph{`#1'}}
\usepackage{comment}
\usepackage{lipsum}
\usepackage{animate}
\usepackage{etoc}
\usepackage{enumitem}

\newenvironment{thebibliographyRenamed}[2]{%
  \renewcommand{\refname}{#2}%
  \begin{thebibliography}{#1}%
  }{\end{thebibliography}%
}

\newenvironment{myquoteenv}[1]%
  {\list{}{\leftmargin=#1\rightmargin=#1}\item[]}%
  {\endlist}

\newcommand{\teachcommander}{\textit{Commander}}
\newcommand{\teachfollower}{\textit{Follower}}

\newcommand{\allenai}{1}
\newcommand{\apple}{2}
\newcommand{\cmu}{3}
\newcommand{\fbk}{4}
\newcommand{\gatech}{5}
\newcommand{\google}{6}
\newcommand{\intel}{7}
\newcommand{\meta}{8}
\newcommand{\nvidia}{9}
\newcommand{\mitN}{10}
\newcommand{\mitibm}{11}
\newcommand{\osu}{12}
\newcommand{\sfu}{13}
\newcommand{\stanford}{14}
\newcommand{\umass}{15}
\newcommand{\padova}{16}
\newcommand{\uw}{17}
\newcommand{\usc}{18}
\newcommand{\utAustin}{19}
\newcommand{\qut}{20}

\title{Retrospectives on the Embodied AI Workshop}

\author{
\footnotesize
\textbf{Matt Deitke}$^{\allenai,\uw}$, \textbf{Dhruv Batra}$^{\gatech,\meta}$, \textbf{Yonatan Bisk}$^{\cmu}$, \textbf{Tommaso Campari}$^{\fbk, \padova}$, \textbf{Angel X. Chang}$^{\sfu}$, \textbf{Devendra Singh Chaplot}$^{\meta}$,\\
\footnotesize
\textbf{Changan Chen}$^{\utAustin}$, \textbf{Claudia P\'{e}rez-D'Arpino}$^{\nvidia}$, \textbf{Kiana Ehsani}$^{\allenai}$, \textbf{Ali Farhadi}$^{\apple,\uw}$, \textbf{Li Fei-Fei}$^{\stanford}$, \textbf{Anthony Francis}$^{\google}$, \textbf{Chuang Gan}$^{\mitibm,\umass}$,\\
\footnotesize
\textbf{Kristen Grauman}$^{\utAustin,\meta}$, \textbf{David Hall}$^{\qut}$, \textbf{Winson Han}$^{\allenai}$, \textbf{Unnat Jain}$^{\meta}$, \textbf{Aniruddha Kembhavi}$^{\allenai,\uw}$, \textbf{Jacob Krantz}$^{\osu}$, \textbf{Stefan Lee}$^{\osu}$, \textbf{Chengshu Li}$^{\stanford}$,\\
\footnotesize
\textbf{Sagnik Majumder}$^{\utAustin}$, \textbf{Oleksandr Maksymets}$^{\meta}$, \textbf{Roberto Martín-Martín}$^{\utAustin}$, \textbf{Roozbeh Mottaghi}$^{\meta,\uw}$, \textbf{Sonia Raychaudhuri}$^{\sfu}$,\\
\footnotesize
\textbf{Mike Roberts}$^{\intel}$, \textbf{Silvio Savarese}$^{\stanford}$, \textbf{Manolis Savva}$^{\sfu}$, \textbf{Mohit Shridhar}$^{\uw}$, \textbf{Niko S\"{u}nderhauf}$^{\qut}$, \textbf{Andrew Szot}$^{\gatech}$, \textbf{Ben Talbot}$^{\qut}$,\\
\footnotesize
\textbf{Joshua B. Tenenbaum}$^{\mitN}$, \textbf{Jesse Thomason}$^{\usc}$, \textbf{Alexander Toshev}$^{\apple}$, \textbf{Joanne Truong}$^{\gatech}$, \textbf{Luca Weihs}$^{\allenai}$, \textbf{Jiajun Wu}$^{\stanford}$\\
\footnotesize
$^{\allenai}$Allen Institute for AI, $^{\apple}$Apple, $^{\cmu}$Carnegie Mellon University, $^{\fbk}$FBK, $^{\gatech}$Georgia Tech, $^{\google}$Google, $^{\intel}$Intel Labs, $^{\meta}$Meta AI, $^{\nvidia}$NVIDIA, $^{\mitN}$MIT,\\
\footnotesize
$^{\mitibm}$MIT-IBM Watson AI Lab, $^{\osu}$Oregon State University, $^{\sfu}$Simon Fraser University, $^{\stanford}$Stanford University, $^{\umass}$UMass Amherst,\\
\footnotesize
$^{\padova}$University of Padova, $^{\uw}$University of Washington, $^{\usc}$University of Southern California, $^{\utAustin}$UT Austin, $^{\qut}$QUT Centre for Robotics
}

\begin{document}

\maketitle

\begin{abstract}
    We present a retrospective on the state of Embodied AI research. Our analysis focuses on 13 challenges presented at the Embodied AI Workshop at CVPR. These challenges are grouped into three themes: (1) visual navigation, (2) rearrangement, and (3) embodied vision-and-language. We discuss the dominant datasets within each theme, evaluation metrics for the challenges, and the performance of state-of-the-art models. We highlight commonalities between top approaches to the challenges and identify potential future directions for Embodied AI research.
\end{abstract}

\begin{figure*}[t!]
    \centering
    \includegraphics[width=1\textwidth]{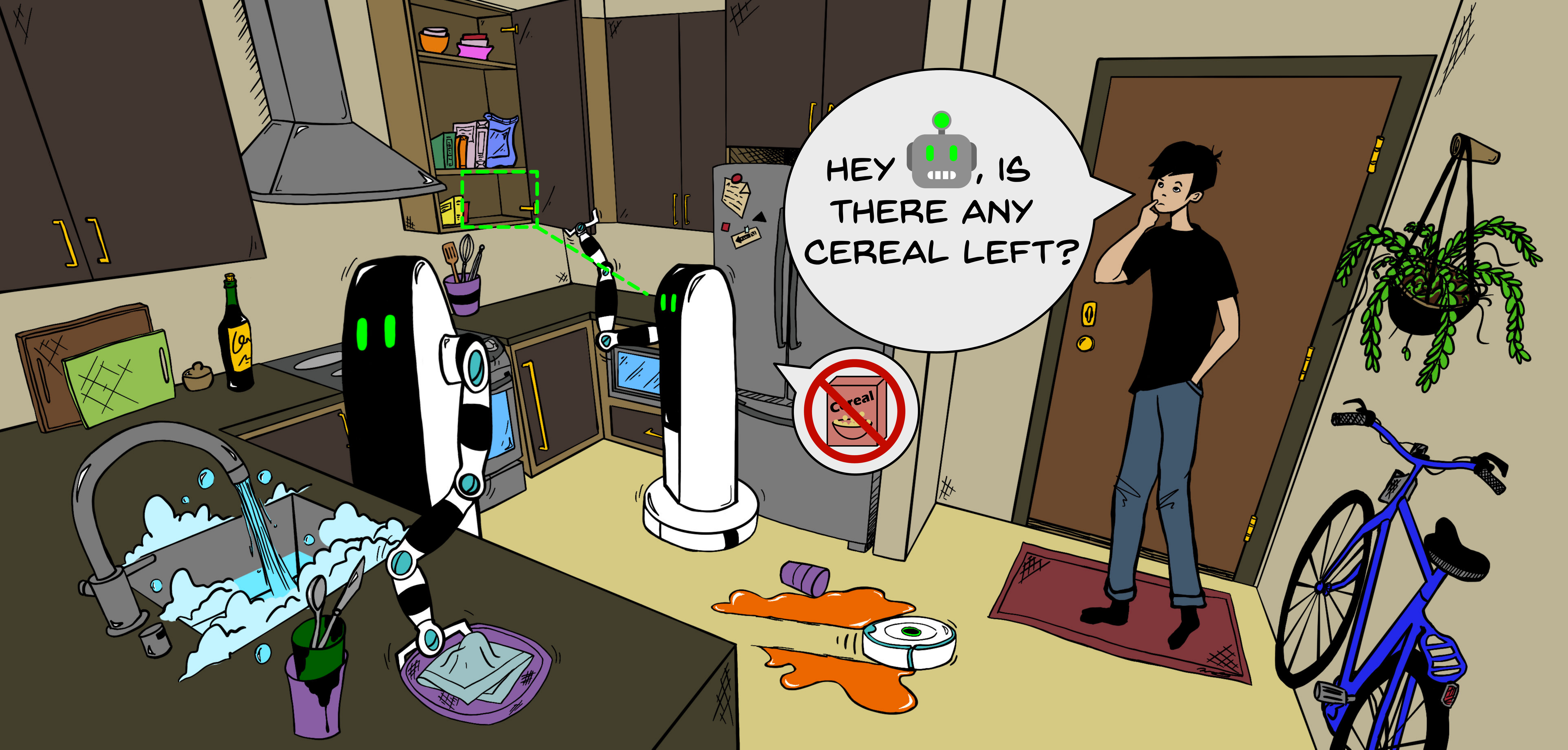}
    \caption{An illustration of a scenario depicting many tasks of interest to researchers in Embodied AI. Here, we have multiple robots operating in a kitchen environment, with a human asking one of the robots if there is any cereal left, while the other one cleans the dishes. The robots must use their navigation, manipulation, and reasoning skills to answer and achieve tasks in the environment.}
\end{figure*}

\section{Introduction}

Within the last decade, advances in deep learning, coupled with the creation of massive datasets and high-capacity models, have resulted in remarkable progress in computer vision, audio, NLP, and the broader field of AI. This progress has enabled models to obtain superhuman performance on a wide variety of passive tasks (\eg image classification). However, this progress has also enabled a paradigm shift towards  embodied agents (\eg robots) which learn, through interaction and exploration, to creatively solve challenging tasks within their environments. The field of embodied AI focuses on how intelligence emerges from an agent's interactions with its environment. An interaction in the environment involves an agent taking an action that affects its future state. For instance, the agent may perform navigation actions to move around the environment or take manipulation actions to open or pick up objects within reach. Embodied AI is a focus of a growing collection of researchers and research challenges.

Consider asking a robot to \myquote{Clean my room} or \myquote{Drive me to my favorite restaurant}. To succeed at these tasks in the real world, the robots need skills like \textit{visual perception} (to recognize scenes and objects), \textit{audio perception} (to receive the speech spoken by the human), \textit{language understanding} (to translate questions and instructions into actions), \textit{memory} (to recall how items should be arranged or to recall previously encountered situations), \textit{physical intuition} (to understand how to interact with other objects), \textit{multi-agent reasoning} (to predict and interact with other agents), and \textit{navigation} (to safely move through the environment). The study of embodied agents both provides a challenging testbed for building intelligent systems and tries to understand how intelligence emerges through interaction with an environment. As such, it involves many disciplines, such as computer vision, natural language processing, acoustic learning, reinforcement learning, developmental psychology, cognitive science, neuroscience, and robotics.

In this paper, we present a retrospective on the state of embodied AI, focusing on the challenges highlighted at the 2020--2022 CVPR embodied AI workshops. The challenges presented in the workshop have focused on benchmarking progress in navigation, rearrangement, and embodied vision-and-language. The navigation challenges include Habitat PointNav~\cite{habitat2020sim2real} and ObjectNav~\cite{batra2020objectnav}, Interactive and Social Navigation with iGibson~\cite{xia2020interactive}, RoboTHOR ObjectNav~\cite{deitke2020robothor}, MultiON~\cite{wani2020multion}, RVSU Semantic SLAM~\cite{hall2020robotic}, and Audio-Visual Navigation with SoundSpaces~\cite{chen_soundspaces_2020}; rearrangement challenges include AI2-THOR Rearrangement~\cite{weihs2021rearrangement}, TDW-Transport~\cite{gan2022threedworld}, and RVSU Scene Change Detection~\cite{hall2020robotic}; and embodied vision-and-language challenges include RxR-Habitat~\cite{rxr}, ALFRED~\cite{ALFRED20}, and TEACh~\cite{teach}. We discuss the setup of each challenge and its state-of-the-art performance, analyze common approaches between winning entries across the challenges, and conclude with a discussion of promising future directions in the field.

\begin{figure*}[ht]
    \centering
    \includegraphics[width=\textwidth]{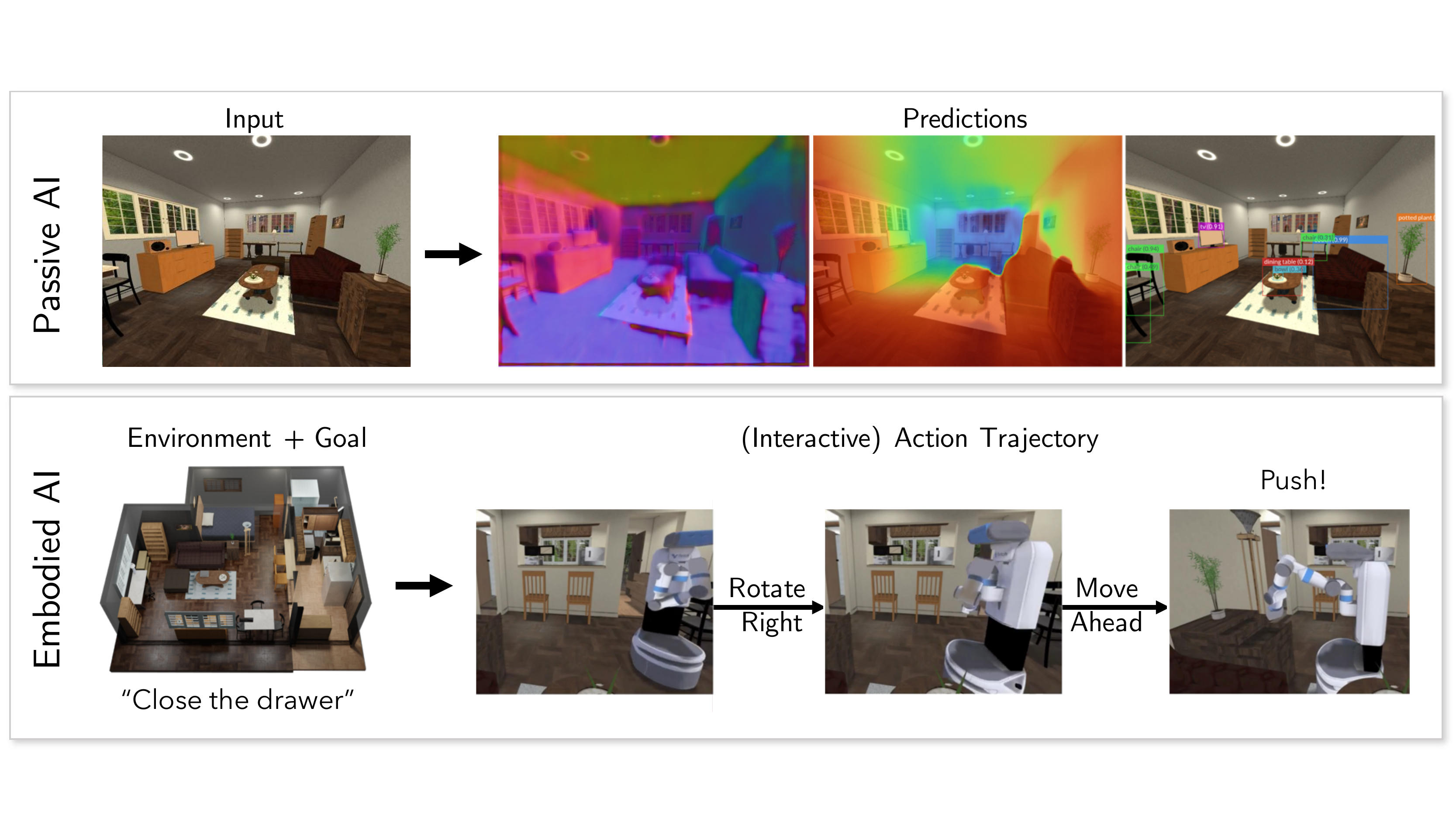}
    \vspace{-12mm}
    \caption{Passive AI tasks are based on predictions over independent samples of the world, such as images collected without a closed loop with a decision-making agent. In contrast, embodied AI tasks include an active artificial agent, such as a robot, that must perceive and interact with the environment purposely to achieve its goals, including in unstructured or even uncooperative settings. 
    Enabled by the progress in computer vision and robotics, embodied AI represents the next frontier of challenges to study and benchmark intelligent models and algorithms for the physical world.}
    \label{fig:passive-vs-embodied-ai}
\end{figure*}

\section{What is Embodied AI?}

\emph{Embodied AI} studies artificial systems that express intelligent behavior through bodies interacting with their environments. The first generation of embodied AI researchers focused on robotic embodiments \cite{pfeifer2004embodied}, arguing that robots need to interact with their noisy environments with a rich set of sensors and effectors, creating high-bandwidth interaction that breaks the fundamental assumptions of clean inputs, clean outputs, and static world states required by \textit{classical AI} approaches \cite{wilkins1988practical}. More recent embodied AI research has been empowered by rich simulation frameworks, often derived from scans of real buildings and models of real robots, to recreate environments more closely resembling the real world than those previously available. These environments have enabled both discoveries about the properties of intelligence \cite{partsey2022mapping} and systems which show excellent sim-to-real transfer \cite{8968053,DBLP:journals/corr/abs-1804-10332}.

Abstracting away from real or simulated embodiments, embodied AI can be defined as the study of intelligent agents that can
\begin{inparaitem}[]
    \item \emph{see} (or more generally perceive their environment through vision, audition, or other senses),
    \item \emph{talk} (\ie hold a natural language dialog grounded in the environment),
    \item \emph{listen} (\ie understand and react to audio input anywhere in a scene.),
    \item \emph{act} (\ie navigate their environment and interact with it to accomplish goals), and 
    \item \emph{reason} (\ie consider the long-term consequences of their actions).
\end{inparaitem}
Embodied AI focuses on tasks which break the clean input/output formalism of passive tasks such as object classification and speech understanding, and require agents to interact with - and sometimes even modify - their environments over time (Fig. \ref{fig:passive-vs-embodied-ai}).
Furthermore, embodied AI environments generally violate the clean dynamics of structured environments such as games and assembly lines, and require agents to cope with noisy sensors, effectors, dynamics, and other agents, which creates unpredictable outcomes.

\paragraph{Why is Embodiment Important?}
Embodied AI can be viewed as a reaction against extreme forms of the \emph{mind-body duality} in philosophy, which some perceive to view intelligence as a purely mental phenomenon. The mind-body problem has faced philosophers and scientists for millennia \cite{crane2012history}: humans are simultaneously ``physical agents'' with mass, volume and other bodily properties, and at the same time ``mental agents'' that think, perceive, and reason in a conceptual domain which seems to lack physical embodiment. Some scholars argue in favor of a strict mind-body duality in which intelligence is a purely mental quality only loosely connected to bodily experience \cite{ryle2009concept}. Other scholars, across philosophy, psychology, cognitive science and artificial intelligence, have challenged this mind-body duality, arguing that intelligence is intrinsically connected to embodiment in bodily experience, and that separating them has distorting effects on research \cite{brooks1990elephants, paul2021extended, varela2017embodied, mehta2011mind, ryle2009concept}.

The history of research in artificial intelligence has mirrored this debate over mind and body, focusing first on computational solutions for symbolic problems which appear hard to humans, a strategy often called GOFAI ("Good Old Fashioned AI", \cite{boden20144,mcdermott2015gofai}). The computational theory of mind argued that if intelligence was reasoning operations in the mind, computers performing similar computations could also be intelligent \cite{piccinini2004first,Sclar2022}. %
Purely symbolic artificial intelligence were often disconnected from the physical world, requiring symbolic representations as input, creating problems with grounding symbols in perception \cite{harnad1990symbol,steels2008symbol} and often leading to brittleness \cite{mccarthy2007here,lohn2020estimating,cummings2020surprising}. However, symbolic reasoning problems themselves often proved to be relatively easy, whereas the physical problems of perceiving the environment or acting in it were actually the most challenging: what is unconscious for humans often requires surprising intelligence, often known as Moravec's Paradox \cite{goldberg2015robotics,agrawal2010study}. 
Some researchers challenged this approach, arguing that for machines to be intelligent, they must interact with noisy environments via rich sets of sensors and effectors, creating high-bandwidth interactions that break the assumption of clean inputs and outputs and discrete states required by \textit{classical AI} \cite{wilkins1988practical}; %
these ideas were echoed by roboticists already concerned with connecting sensors and actuators more directly \cite{arkin1998behavior,brooks1990elephants,moravec2000ripples}.
Much as neural network concepts hibernated through several AI winters before enjoying a renaissance, embodied AI ideas have now been revived by new interest from fields such as computer vision, machine learning and robotics - often in combination with neural network ideas.
New generations of artificial neural networks are now able to digest raw sensor signals, generate commands to actuators, and autonomously learn problem representations, linking "classical AI" tasks to embodied setups.

Thus, embodied AI is more than just the study of agents that are active and situated in their environments: it is an exploration of the properties of intelligence. Embodied AI research has demonstrated that intelligent systems that perform well at embodied tasks often look different than their passive counterparts \cite{fu2022coupling} - but, conversely, that highly performing passive AI tasks can often contribute greatly to embodied systems as components \cite{shridhar2022cliport}. Furthermore, the control over embodied agents provided by modern simulators and deep learning libraries enables ablation studies that reveal fine-grained details about the properties needed for individual embodied tasks \cite{partsey2022mapping}.

\paragraph{What is \emph{not} Embodied AI?}

Embodied AI overlaps with many other fields, including robotics, computer vision, machine learning, artificial intelligence, and simulation.
However, there are differences in focus which make embodied AI a research area in its own right.

All \textit{robotic} systems are embodied; however, not all embodied systems are robots (e.g., AR glasses), and robotics requires a great deal of work beyond purely trying to make systems intelligent. Embodied AI also includes work that focuses on exploring the properties of intelligence in realistic environments while abstracting some of the details of low-level control.
For example, the ALFRED~\cite{ALFRED20} benchmark uses simulation to abstract away low-level robotic manipulation (\eg moving a gripper to grasp an object) to focus on high-level task planning. Here, the agent is tasked with completing a natural language instruction, such as \textit{rinse the egg to put it in the microwave}, and it can open or pickup an object by issuing a high-level \textit{Open} or \textit{Pickup} action that succeeds if the agent is looking at the object and is sufficiently close to it.
Additionally, \cite{partsey2022mapping} provides an example of studying properties of intelligence, where they attempt to answer whether mapping is strictly required for a form of robotic navigation.
Conversely, robotics includes work that focuses directly on the aspects of the real world, such as low-level control, real-time response, or sensor processing.

\textit{Computer vision} has contributed greatly to embodied AI research; however, computer vision is a vast field, much of which is focused purely on improving performance on passive AI tasks such as classification, segmentation, and image transformation. Conversely, embodied AI research often explores problems that require other modalities with or without vision, such as navigation with sound \cite{chen_soundspaces_2020} or pure LiDAR images.

\textit{Machine learning} is one of the most commonly used techniques for building embodied agents. However, machine learning is a vast field encompassing primarily passive tasks, and most embodied AI tasks are formulated in such a way that they are learning agnostic. For example, the iGibson 2020 challenge \cite{shen2020igibson} allowed training in simulated environments but deployment in holdout environments in both real and simulation; nothing required the solutions to use a learned approach as opposed to a classical navigation stack (though learned approaches were the ones deployed).

\textit{Artificial intelligence} is written into the name of embodied AI, but the field of embodied AI was created to address the perceived limitations of classical artificial intelligence \cite{pfeifer2004embodied}, and much of artificial intelligence is focused on problems like causal reasoning or automated programming which are hard enough without introducing the messiness of real embodiments. More recently, techniques from more traditional artificial intelligence domains like natural language understanding have been applied to embodied problems with great success \cite{ahn2022can}.

\textit{Simulation} and embodied AI are intimately intertwined; while simulations of real-world systems go far beyond the topics of robotics, and the first generation of embodied AI focused on robotic embodiments \cite{pfeifer2004embodied}, much of modern embodied AI research has expanded to simulated benchmarks, emulating or even scanned from real environments, which provide challenging problems for traditional AI approaches, with or without physical embodiments. Despite not starting with robots, systems that have resulted from this work have nevertheless found success in real-world environments \cite{8968053,DBLP:journals/corr/abs-1804-10332}, providing hope that simulated benchmarks will prove a fruitful way to develop more capable real-world intelligent systems.

\paragraph{Why focus on real-world environments?}
Many researchers are exploring intelligence in areas such as image recognition or natural language understanding where at first blush interaction with an environment appears not to be required. Genuine discoveries about intelligent systems appear to have been made here, such as the role of convolutions in image processing and the role of recurrent networks and attention in language processing. So a reasonable question is, why do we need to focus on interactive and realistic (if not real-world) environments if we want to understand intelligence?

Focusing on interactive environments is important because each new modality of intelligence we consider - classification, image processing, natural language understanding, and so on - has required new architectures for learning systems \cite{goodfellow2016deep}, \cite{chollet2021deep}. Interacting with an environment over time requires the techniques of reinforcement learning. Deep reinforcement learning has made massive strides in creating learning systems for synthetic environments, including traditional board games, Atari games, and even environments with simulated physics such as the Mujoco environments.

However, embodied AI research focuses on environments that are either more realistic \cite{habitatchallenge2022} or which require actual deployments in the real world \cite{habitat2020sim2real,shen2020igibson}). This shift in emphasis has two primary reasons. First, many embodied AI researchers
believe that the challenges of realistic environments are critical for developing systems that can be deployed in the real world. Second, many embodied
AI researchers believe that there are genuine discoveries to be made about the properties of intelligence needed to handle real world environments that can only be made by attempting to solve problems in environments that are as close to the real world as is feasible at this time.

\section{Challenge Details}

In this section, we discuss the 13 challenges present at our Embodied AI Workshop between 2020--2022. The challenges are partitioned into navigation challenges, rearrangement challenges, and embodied vision-and-language challenges. Most challenges present a distinctive tasks, metrics and training datasets, though many challenges share similar observation spaces, action spaces, and environments.

\subsection{Navigation Challenges}

Our workshop has featured a number of challenges relating to embodied visual navigation. At a high-level, the tasks consist of an agent operating in a simulated 3D environment (\eg a household), where its goal is to move to some target. For each task, the agent has access to an egocentric camera and observes the environment from a first-person's perspective. The agent must learn to navigate the environment from its visual observations.

The challenges primarily differ based on how the target is encoded (\eg ObjectGoal, PointGoal, AudioGoal), how the agent is expected to interact with the environment (\eg static navigation, interactive navigation, social navigation), the training and evaluation scenes (\eg 3D scans, video-game environments, the real world), the observation space (\eg RGB vs. RGB-D, whether to provide localization information), and the action space (\eg outputting discrete high-level actions or continuous joint movement actions).

\noindent
\begin{minipage}[l]{0.46\textwidth}
    \vspace{0.1in}
    \centering
    \includegraphics[width=\textwidth]{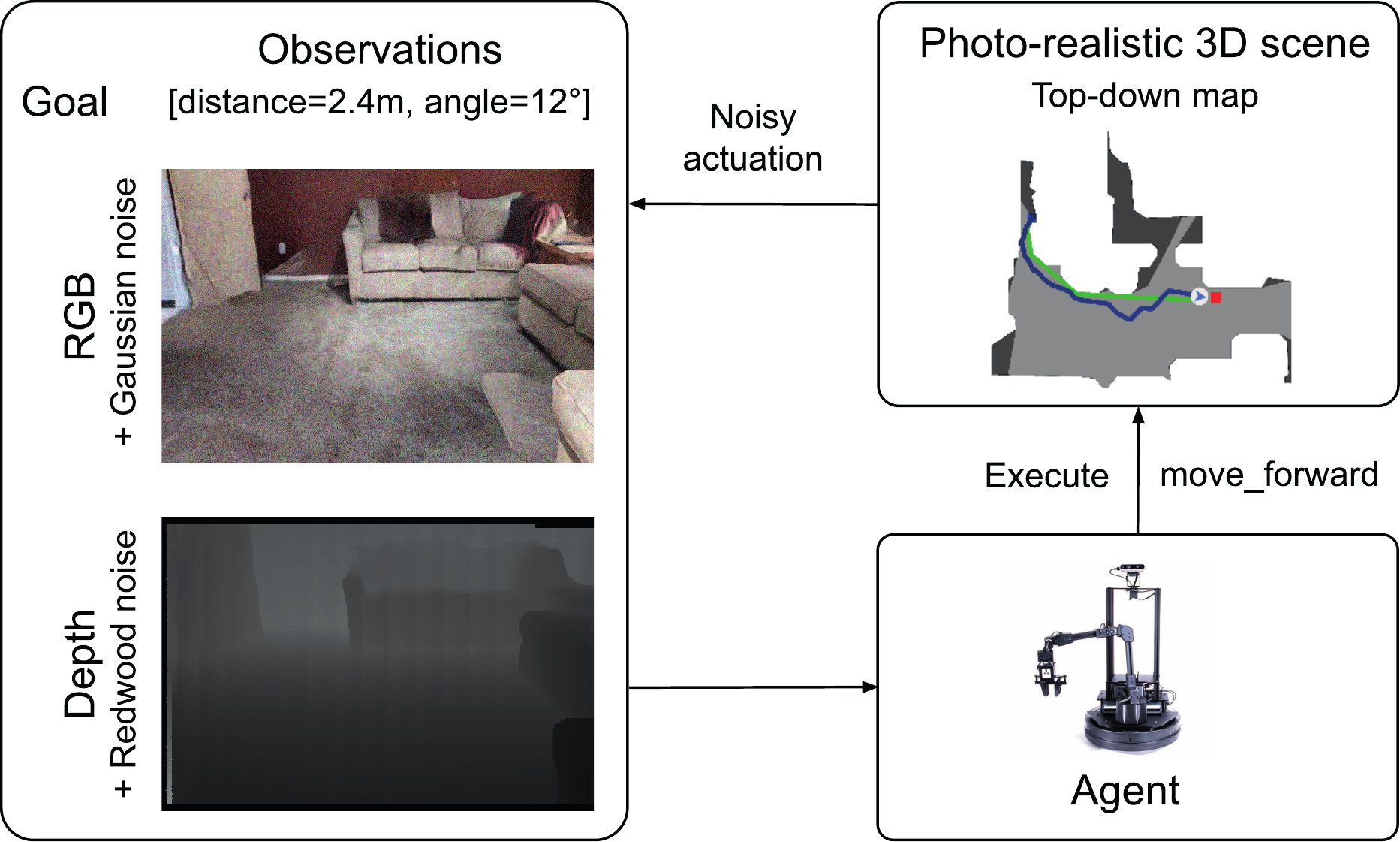}
    \vspace{-0.1in}
    \captionsetup{type=figure}
    \captionof{figure}{The \emph{PointNav} task requires an agent to navigate to a goal coordinate in a novel environment (potentially with noisy sensory inputs), without access to a pre-built map of the environment.}
    \vspace{-0.1in}
\end{minipage}

\subsubsection{PointNav}

In PointNav, the agent's goal is to navigate to target coordinates in a novel environment that are relative to its starting location (\eg navigate 5m north, 3m west relative to its starting pose), without access to a pre-built map of the environment. The agent has access to egocentric sensory inputs (RGB images, depth images, or both), and an egomotion sensor (sometimes referred to as GPS+Compass sensor) for localization. The action space for the robot consists of: \emph{Move Forward 0.25m}, \emph{Rotate Right $30^\circ$}, \emph{Rotate Left $30^\circ$}, and \emph{Done}. An episode is considered successful if the agent issues the \emph{Done} command within 0.2 meters of the goal and within 500 maximum steps. The agent is evaluated using the Success Rate (SR) and "Success
weighted by Path Length" (SPL) \cite{anderson_arxiv18} metrics, which measures the success and efficiency of the path taken by the agent. For training and evaluation, challenge participants use the train and val splits from the Gibson 3D dataset \cite{zamir_cvpr18}. 

In 2019, AI Habitat hosted its first challenge on PointNav.\footnote{\url{https://aihabitat.org/challenge/2019/}} The winning submission \cite{chaplot2020learning} utilized a combination of classical and learning-based methods, and achieved a high test SPL of 0.948 in the RGB-D track, and 0.805 in the RGB track. In 2020 and 2021, the PointNav challenge was modified to emphasize increased realism and on sim2real predictivity (the ability to predict performance on a real robot from its performance in simulation) based on findings from Kadian et al. \cite{habitatsim2real20ral}. Specifically, the challenge (PointNav-v2) introduced (1) no GPS+Compass sensor, (2) noisy actuation and sensing, (3) collision dynamics and `sliding’, and (4) minor changes to the robot embodiment/size, camera resolution, height to better match the LoCoBot robot. These changes proved to be much more challenging, with the winning submission in 2020 \cite{ramakrishnan2020occant} achieving a SPL of 0.21 and SR of 0.28. In 2021, there was a major breakthrough with a 3$\times$ performance improvement over the winners in 2020; the winning submission achieved a SPL of 0.74 and SR of 0.96 \cite{habitat2020sim2real}. Since an agent with perfect GPS + Compass sensors in this PointNav-v2 setting can only achieve a maximum of 0.76 SPL and 0.99 SR, the PointNav-v2 challenge was considered solved, and discontinued in future years.

\subsubsection{Interactive and Social PointNav}

\begin{figure}[ht!]
     \centering
     \begin{subfigure}[b]{0.23\textwidth}
         \centering
         \includegraphics[width=1\textwidth]{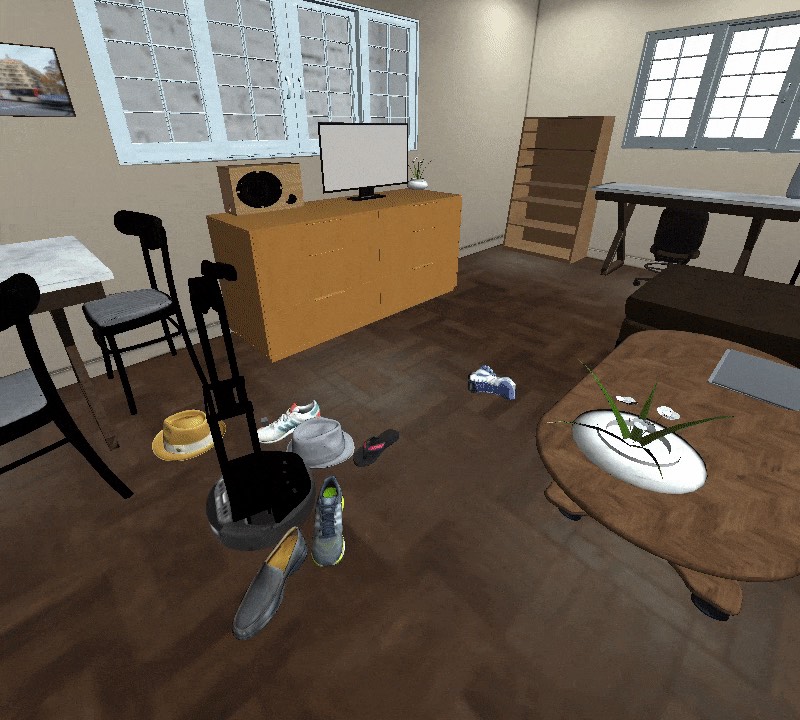}
         \caption{Interactive Navigation}
     \end{subfigure}
     \hfill
     \begin{subfigure}[b]{0.23\textwidth}
         \centering
         \includegraphics[width=1\textwidth]{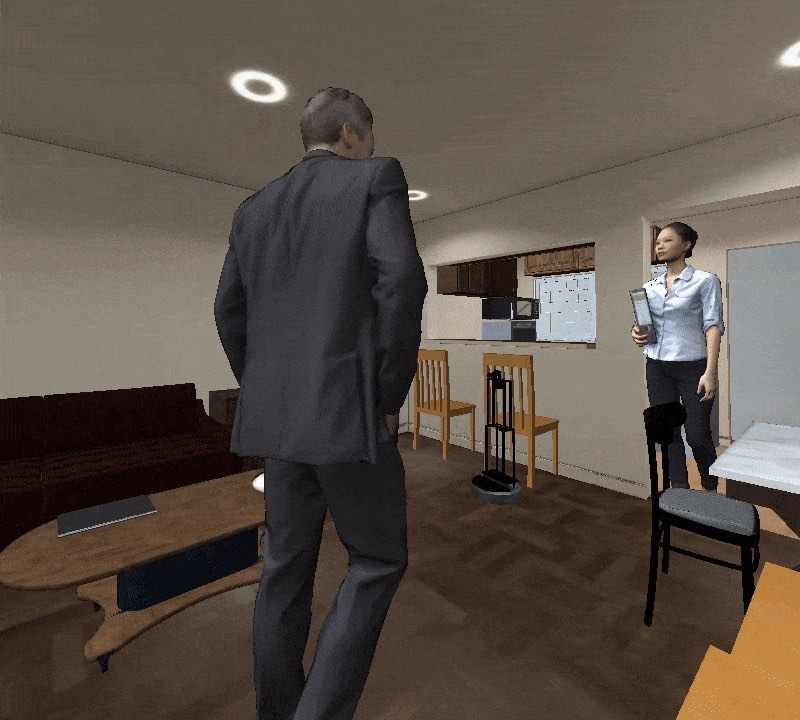}
         \caption{Social Navigation}
     \end{subfigure}
    \caption{\emph{Interactive Navigation} (left) requires the agent to push aside small obstacles (\eg shoes, boxes) whereas \emph{Social Navigation} (right) requires the agent to navigate among pedestrians and respect their personal space.}
    \label{fig:iGibsonChallenges}
\end{figure}

In Interactive and Social Navigation, the agent is required to reach a PointGoal in dynamic environments that contain dynamic objects (furniture, clutter, etc) or dynamic agents (pedestrians). Although robot navigation achieves remarkable success in static, structured environments like warehouses, it still remains a challenging research question in dynamic environments like homes and offices. In 2020 and 2021, the Stanford Vision and Learning Lab in collaboration with Robotics@Google hosted challenges on Interactive and Social (Dynamic) Navigation\footnote{\url{https://svl.stanford.edu/igibson/challenge2021.html}}. These challenges used the simulation environment iGibson~\cite{shen2020igibson, li2021igibson} with a number of realistic indoor scenes, as illustrated in Fig.~\ref{fig:iGibsonChallenges}. The 2020 Challenge\footnote{\url{https://svl.stanford.edu/igibson/challenge2020.html}} also featured a Sim2Real component where the participants trained their policies in the iGibson simulation environment and deployed in the real world.

In \textit{Interactive Navigation}, we challenge the notion that navigating agents are to avoid collision at any cost. We argue for the contrary -- in clutter-filled real environments, such as homes, an agent will have to interact and push away objects to achieve meaningful navigation. Note that all objects in the scenes are assigned realistic physical weight and are  interactable. As in the real world, while some objects are light and movable by the robot, others are not. Along with the furniture objects originally in the scenes, additional objects (e.g. shoes and toys) from the Google Scanned Objects dataset~\cite{downs2022google} are added to simulate real-world clutter. The performance of the agent is evaluated using a novel Interactive Navigation Score (INS)~\cite{xia2020interactive} that measures both navigation success as well as the level of disturbance to the scene an agent has caused along the way.

In \textit{Social Navigation}, the agent navigates among walking humans in a home environment. The humans in the scene move towards randomly sampled locations, and their 2D trajectories are simulated using the model of Optimal Reciprocal Collision Avoidance (ORCA)~\cite{berg2011reciprocal} integrated in iGibson~\cite{shen2020igibson, li2021igibson, darpino2021socialnav}. The agent shall avoid collisions or proximity to pedestrians beyond a threshold (distance <0.3 meter) to avoid episode termination. It should also maintain a comfortable distance to pedestrians (distance <0.5 meter), beyond which the score is penalized but episodes are not terminated. Social Navigation Score (SNS), which is the average of STL (Success weighted by Time Length) and PSC (Personal Space Compliance), is used to evaluate performance of the agent.

The agent takes in the current RGB-D images, the target coordinates in its local frame, and current velocities as observations, and outputs a continuous twist command (desired linear and angular velocities) as actions. The dataset includes eight training scenes, two validation scenes and five testing scenes. All scenes are fully interactive.

In the 2020 edition we saw 4 submissions while in the subsequent 2021 edition we had 6 submissions. The current state-of-the-art learning based methods achieved some level of success for Interactive and Social Navigation tasks (around $0.5$ INS and $0.45$ SNS), but they are still far from being solved. In both competitions participants improved over navigation success rate while keeping environment disturbance relatively constant. The common failure cases include the agent being too conservative and not being able to clear the obstacles in time, and the agent being too aggressive and colliding with the other moving pedestrians. 

One of the challenges for the Social Nav part was the difficulty in simulating the trajectories of the human agents, including reactivity and interaction between agents. Often times, getting to the goal requires negotiation of the space or the agent would require to go over the desired personal space threshold; or the simulated human agents behave erratically due to limitations on the behavior models and the space constraints. For future editions, we are to emphasize on the importance of high fidelity simulation of navigation with human-like behaviors.

For the Sim2Real component of the 2020 Challenge, a significant performance drop was observed during the Sim2Real transfer, due to the reality gap in visual sensor readings, dynamics (\eg motor actuation), and 3D modeling (\eg soft carpets). More analysis of the takeaways can be found in the iGibson Challenge 2020\footnote{\url{https://www.youtube.com/watch?v=0BvUSjcc0jw}} and 2021\footnote{\url{https://www.youtube.com/watch?v=1uSsds7HSrQ}} videos, along with the winning entry paper~\cite{yokoyama2021benchmarking}.

\subsubsection{ObjectNav}

In ObjectNav, the agent is tasked with navigating to one of a set of target object types (\eg navigate to the bed) given ego-centric sensory inputs. The sensory input can be an RGB image, a depth image, or combination of both. At each time step the agent must issue one of the following actions: \emph{Move Forward}, \emph{Rotate Right}, \emph{Rotate Left}, \emph{Look Up}, \emph{Look Down}, and \emph{Done}. The \emph{Move Forward} action moves the agent by 0.25m and the rotate and look actions are performed in $30^\circ$ increments.

Episodes are considered successful if (1) the object is visible in the camera's frame (2) the distance between the agent and the target object is within 1 meter and (3) the agent issues the Done action. The starting location of the agent is a random location in the scene.

Our workshop has held 2 ObjectNav challenges: the RoboTHOR ObjectNav Challenge~\cite{deitke2020robothor} and the Habitat ObjectNav Challenge~\cite{habitatchallenge2022, savva2019habitat}. Both challenges use the mentioned action and observation space, as well as a simulated LoCoBot robotic agent. In comparison:
\begin{itemize}
    \item \textit{Scenes.} The RoboTHOR Challenge\footnote{\url{https://ai2thor.allenai.org/robothor/challenge}} includes 89 room-sized dorm-like scenes. The Habitat 2021 Challenge\footnote{\url{https://aihabitat.org/challenge/2021/}} 90 houses from the Matterport3D dataset~\cite{Chang3DV2017Matterport} and the Habitat 2022 Challenge\footnote{\url{https://aihabitat.org/challenge/2022/}} uses 120 houses from the HM3D Semantics dataset~\cite{ramakrishnan2021habitat}. Both iterations of the Habitat Challenge use scenes collected from real-world scans. In contrast, RoboTHOR scenes were hand-built by 3D artists to be accessible in AI2-THOR \cite{ai2thor} in the Unity game engine. Habitat houses are significantly larger than those in RoboTHOR, often consisting of multiple floors.
    \item \textit{Target Objects.} The RoboTHOR Challenge uses 13 relatively small objects as target object types (\eg Alarm Clock, Basketball, Laptop). The Habitat 2021 Challenge used 21 target objects types and the Habitat 2022 Challenge used 6 target object types. The target object types in both Habitat Challenges typically represent larger objects (\eg Bed, Fireplace, Sofa).
\end{itemize}

For the RoboTHOR Challenge, state-of-the-art is currently held by ProcTHOR~\cite{deitke2022procthor}, which has a test SPL~\cite{anderson_arxiv18} of 0.2884 and a success rate of 65\% on unseen scenes during training. ProcTHOR uses a fairly simple model that embeds images with CLIP, feeds it through a GRU, and uses an actor-critic output optimized with DD-PPO. Its novelty is pre-training on 10K procedurally generated houses (ProcTHOR-10K). It then fine tunes in RoboTHOR. For the Habitat 2022 Challenge, state-of-the art by SPL is also held by ProcTHOR, achieving 0.32 SPL and a success rate of 54\% on unseen scenes. For the Habitat 2022 Challenge, ProcTHOR pre-trains on ProcTHOR-10K and fine-tunes on the HM3D Semantics scenes. When sorting the Habitat 2022 Challenge entries by success rate, imitation learning with Habitat-Web~\cite{ramrakhya2022habitat}, fine-tuned with RL, achieves a state-of-the-art 60\% success rate and an SPL of 0.30 on unseen scenes. Habitat-Web built a web interface to collect human demonstrations of ObjectNav with Amazon Mechanical Turk. It also achieved state-of-the-art in the Habitat 2021 Challenge, with an SPL of 0.146 and a success rate of 34\%.

\noindent
\begin{figure}[ht!]
    \vspace{0.1in}
    \centering
    \textbf{Task:} Find the Bed\\[-0.2in]
    \includegraphics[width=0.475\textwidth]{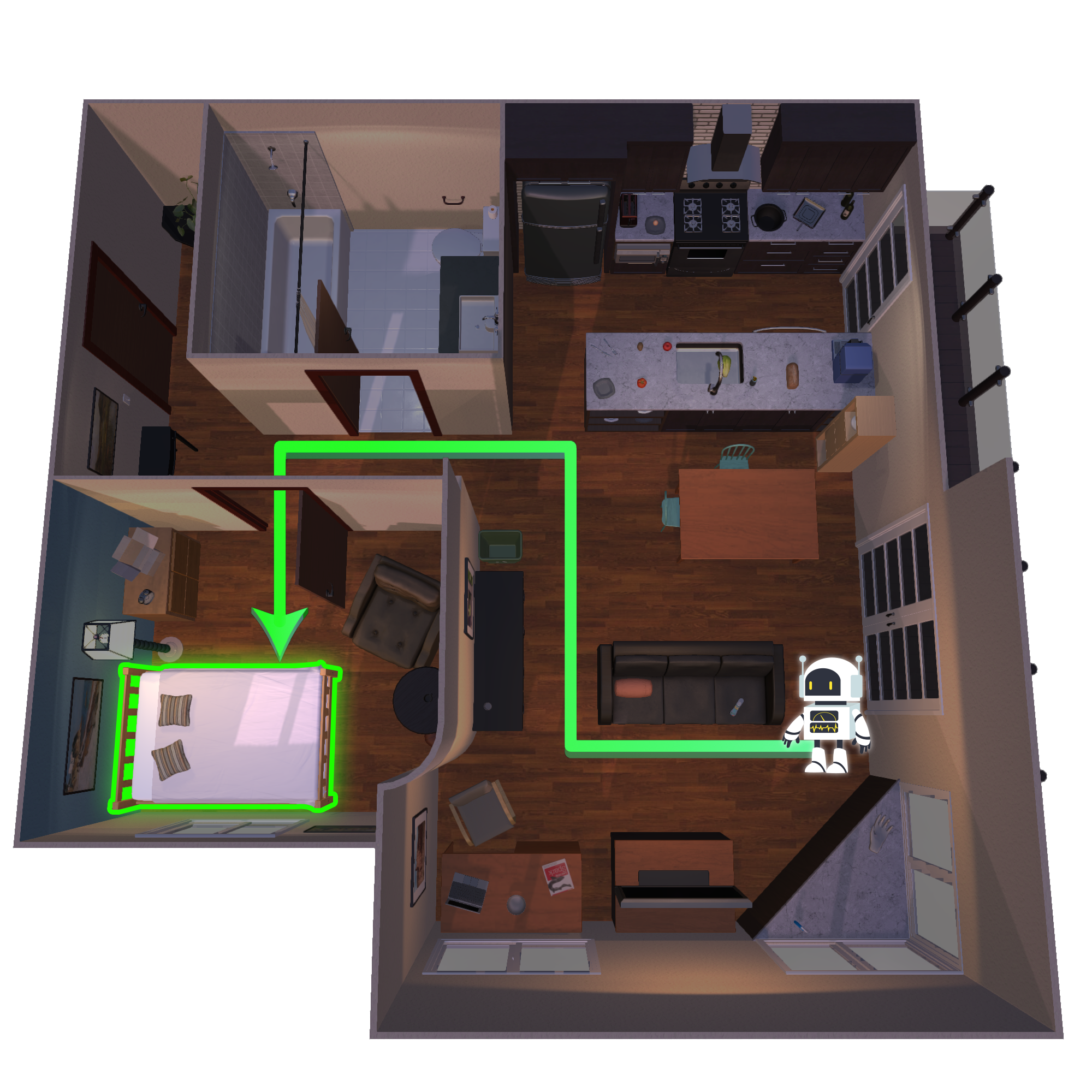}
    \vspace{-0.1in}
    \caption{
    \emph{ObjectNav} tasks the agent with navigating to a given object type in the scene. This example shows the agent tasked with navigating to the \emph{Bed} in the scene. The house is curtousy of the ArchitecTHOR dataset \cite{deitke2022procthor}.
    }
    \vspace{-0.1in}
\end{figure}

\begin{figure}[ht!]
     \centering
     \begin{subfigure}[b]{0.46\textwidth}
         \centering
         \includegraphics[width=1\textwidth]{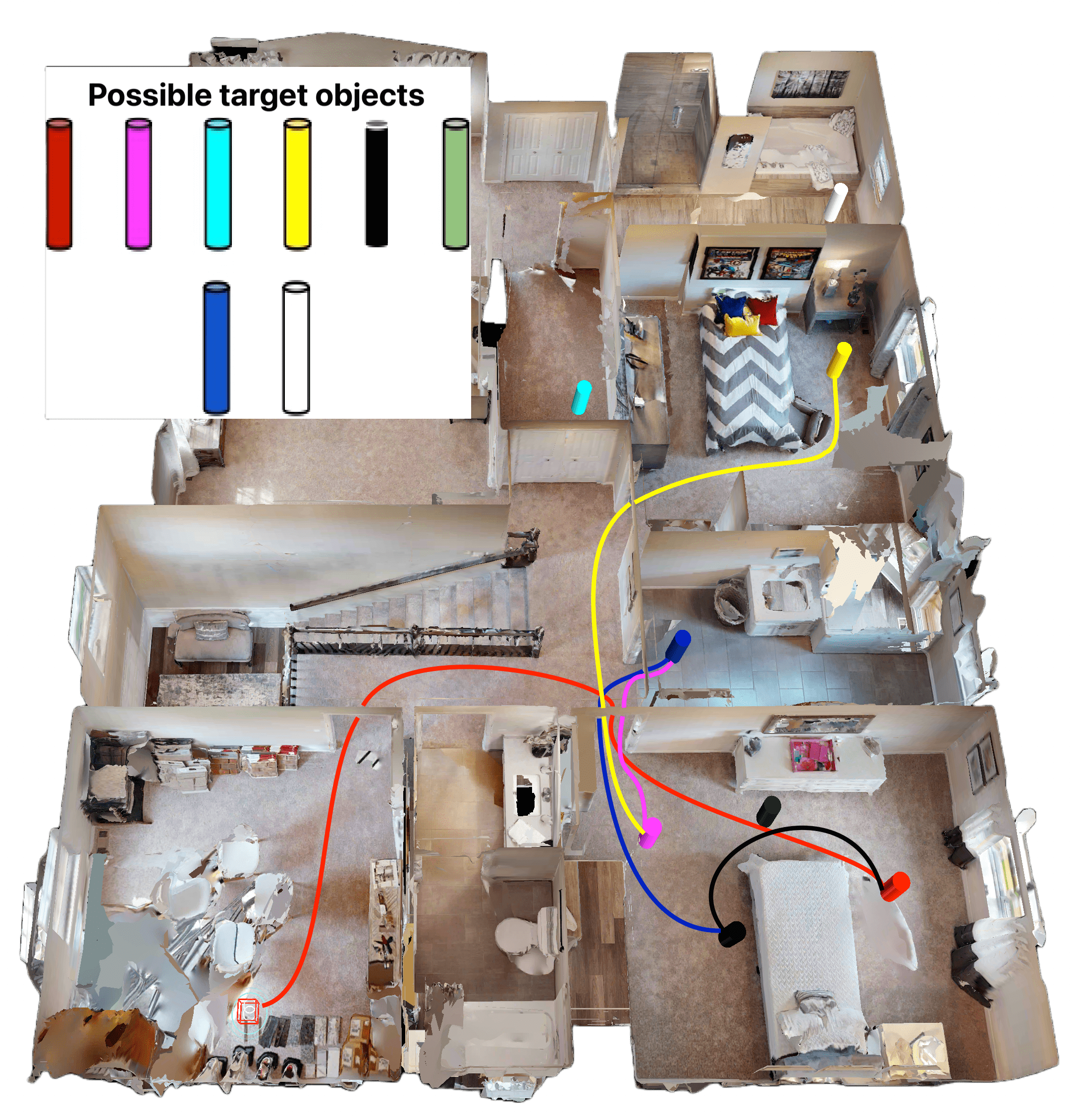}
         \caption{}
         \label{fig:multion_cyl}
     \end{subfigure}
     \hfill
     \begin{subfigure}[b]{0.48\textwidth}
         \centering
         \includegraphics[width=1\textwidth]{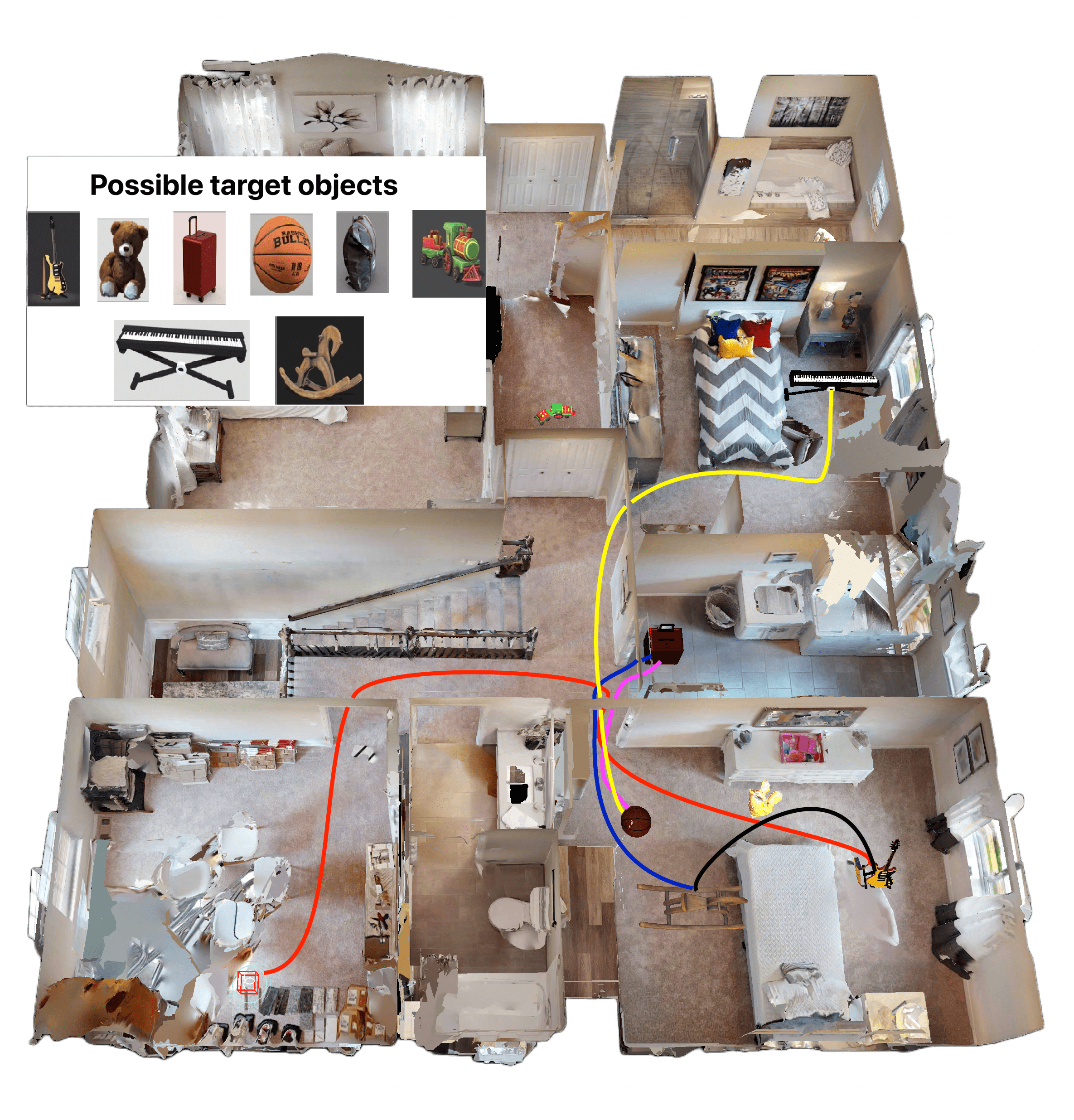}
         \caption{}
         \label{fig:multion_real}
     \end{subfigure}
    \caption{\emph{Multi-ObjectNav}: (a) Top-down visualization of a MultiON episode with 5 target cylinder objects in a particular sequence; (b) Top-down visualization of a MultiON episode with 5 target real objects in a particular sequence.}
\end{figure}

\subsubsection{Multi-ObjectNav}
In Multi-ObjectNav (MultiON)~\cite{wani2020multion}, the agent is initialized at a random starting location in an environment and asked to navigate to an ordered sequence of objects placed within realistic 3D interiors (Figures~\ref{fig:multion_cyl},~\ref{fig:multion_real}). The agent must navigate to each target object in the given sequence and call the \emph{Found} action to signal the object's discovery.
This task is a generalized variant of ObjectNav, whereby the agent must navigate to a sequence of objects rather than a single object.
MultiON explicitly tests the agent's navigation capability in locating previously observed goal objects and is, therefore, a suitable test bed for evaluating memory-based architectures for Embodied AI.

The agent is equipped with an RGB-D camera and a (noiseless) GPS+Compass sensor.
The GPS+Compass sensor provides the agent's current location and orientation relative to its initial location and orientation in the episode.
It is not provided with a map of the environment. The action space comprises of \emph{Move Forward} by 0.25 meters, \textit{Rotate Left} by 30$^\circ$, \textit{Rotate Right} by 30$^\circ$ and \emph{Found}.

The MultiON dataset is created by synthetically adding objects in the Habitat-Matterport 3D (HM3D)~\cite{ramakrishnan_arxiv21} scenes. The objects are either cylinder-shaped or natural-looking (real) objects. As shown in Figure~\ref{fig:multion_cyl},  the cylinder objects are of the same height and radius, with different colors. However, such objects do not appear realistic in the indoor scenes of Matterport houses. Furthermore, detecting the same object with different colors might be easy for the agent to learn. This has led us to include realistic-looking objects that can naturally occur in houses (Figure~\ref{fig:multion_real}). These objects are of varying sizes and shapes and pose a more demanding detection challenge. 
There are 800 HM3D scenes and 8M episodes in the training split, 30 unseen scenes and 1050 episodes in the validation split, and 70 unseen scenes and 1050 episodes in the test split.
The episodes are generated by sampling random navigable points as start and goal locations, such that the locations are on the same floor and a navigable path exists between them. Next, five goal objects are randomly sampled from the set of Cylinder or Real objects to be inserted between the start and the goal, maintaining a minimum pairwise geodesic distance between them to avoid cluttering. Furthermore, to make the task even more realistic and challenging, three distractor objects (which are not goals) are inserted in each episode. The presence of distractors will encourage new agents to distinguish between goal objects and other objects in the environment. An episode is considered successful if the agent is able to reach within 1 meter of every goal in the specified order and generate the \textit{FOUND} action at each goal object. Apart from the standard evaluation metrics used in ObjectNav, such as Success Rate (SR) and Success weighted by path length (SPL)~\cite{anderson_arxiv18}, we additionally use Progress and Progress weighted by path length (PPL) to measure agent performance. The leaderboard for the challenge is based on the PPL metric. MultiON challenge was hosted on evalAI, an open-source platform for evaluating and comparing artificial intelligence methods. The participants implemented their methods in docker images and submitted them to evalAI. The docker images were evaluated on evaluation servers, and the results were uploaded to evalAI.

The MultiON task is similar to ObjectNav, but at the same time, it tries to solve different challenges. Notably, it aims to inject long-term planning capabilities into the agents. In the ObjectNav task, the object detection task takes on a fundamental role. Still, the agent does not have to remember all the objects (and their semantic information) encountered in the past. In MultiON, on the other hand, we assume a more limited part of the detection (e.g., detecting cylinders or a set of a limited number of natural objects). Parallelly, the agent must be able to remember the objects already seen. Thus, this task is more tailored to the real world than ObjectNav. In fact, the agents operate in the same environment for a very long time and, therefore, must be able to remember what has already been seen. For this reason, the approaches developed for MultiON, unlike those for ObjectNav, always add a component that stores the semantic information obtained through exploration.

For the 2021 challenge, a simpler setup was used. The distractors were absent, the objects were only the cylinders, and the dataset was developed on Matterport3D \cite{chang2017matterport3d}. The Proj-Neural model was used as Baseline \cite{wani2020multion}. This model takes advantage of an egocentric map that is used as an input for an end-to-end model that achieved 29\% Progress and 12\% Success. Surprisingly, two models based on mapping and path planning, SgoLAM (64\% progress, 52\% Success) and Memory Augmented SLAM (Mem-SLAM) (57\% Progress, 36\% Success), exceeded the results obtained from the Baseline by a large margin demonstrating that this type of model works well on long-horizon tasks. Instead, the model proposed in \cite{marza2021teaching} won the 2021 challenge, with a progress of 67\% and a success of 55\%. This model is an evolution of Proj-Neural, where three auxiliary tasks were used to inject information about the map and objects into the agent’s internal representation.

In the 2022 challenge, instead, we noticed some similarities between the Baseline method, Mem-SLAM, and the winning entry in the 2022 MultiON challenge, Exploration and Semantic Mapping for Multi Object-Goal Navigation (EXP-MAP). Both the methods are modular, consisting of detection (identifying objects from raw RGB images), Mapping (incrementally building a top-down map of the environment using Depth observations and relative poses), and Planning (navigating to a detected goal object by generating low-level actions) modules. All these models record previously seen objects in some memory (e.g., semantic map of the environment). The EXP-MAP can achieve 70\% Progress and 60\% Success in the Test-Challenge split of the Cylinder objects track of the challenge while achieving 55\% Progress and 40\% Success in the Real objects track. These results show that episodes with natural objects are more challenging to detect than the cylinders.

\noindent
\begin{minipage}[l]{0.46\textwidth}
    \vspace{0.1in}
    \centering
    \includegraphics[width=\textwidth, height=1.75in]{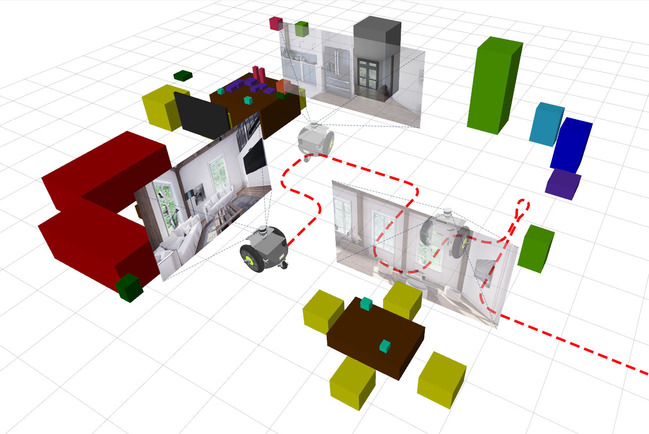}
    \vspace{-0.1in}
    \captionsetup{type=figure}
    \captionof{figure}{In the \emph{RVSU Semantic SLAM} task, an autonomous agent explores environment to create a semantic 3D cuboid map of objects.}
    \vspace{-0.1in}
\end{minipage}

\subsubsection{Navigating to Identify All Objects in a Scene}

The RVSU semantic SLAM challenge tasks participants with exploring a simulation environment to map out all objects of interest therein. 
This challenge asks a robot agent the question, ``what objects are where?'' within the scene.
Robot agents traverse a scene, create an axis aligned 3D cuboid semantic map of the objects within that scene, and are evaluated based on their map's accuracy.
Providing a semantic understanding of objects can assist a robot's ability to interpret attributes of its environment such as knowing how to interact with objects and understanding what type of room it might be in.
This semantic understanding is typically viewed as a semantic simultaneous localization and mapping (SLAM) problem.
The task of semantic SLAM has already seen great investigation using static datasets such as KITTI~\cite{Geiger2013IJRR}, Sun RGBD~\cite{song2015sun} and SceneNet~\cite{McCormac:etal:ICCV2017}.
However, these static datasets ignore the active capabilities of robots and forego searching the physical action space for the actions that best explore and understand an environment.
Addressing this limitation, the RVSU semantic SLAM challenge~\cite{hall2020robotic} helps bridge the gap between passive and active semantic SLAM systems by providing a framework and simulation environments for repeatable, quantitative comparison of both passive and active approaches.

Participation in the challenge is conducted through simulated environments, accessed and controlled using the BenchBot framework~\cite{talbot2020benchbot}.
The environments used are a version of the BenchBot environments for active robotics (BEAR)~\cite{hall2022bear} rendered using the NVIDIA Omniverse Isaac Simulator\footnote{\url{https://developer.nvidia.com/isaac-sim}}.
BEAR provides 25 high-fidelity indoor environments comprising of five base environments with five variations thereof.
Between variations, objects are added and removed, and lighting conditions are changed.
Across environments there are 25 object classes of interest to be mapped within the challenge.
The challenge splits BEAR into 2 base environments for algorithm development and 3 for final testing and evaluation.
The BenchBot framework enables a simulated robot to explore BEAR using either passive or active control  through discretised actions that are pre-defined or actively chosen by the agent respectively.
The action space for robot agents is \textit{MOVE\_NEXT} for passive mode and \textit{MOVE\_DISTANCE} and \textit{MOVE\_ANGLE} for active mode with magnitude of movement being defined by users with a minimum distance of 0.01 m and a minimum angle of 1°.
BenchBot provides the robot agent access to RGB-D camera, laser, and either ground-truth or estimated pose information for the robot immediately after completing any given action.
The progression of passive control with ground-truth pose data, through to active control with estimated pose data is designed to gradually bridge the gap from passive to active semantic SLAM.
The final cuboid map created by the agent within the challenge is evaluated using the new object map quality (OMQ) measure outlined in~\cite{hall2020robotic}.
This evaluation measure considers the quality of every provided object cuboid, in terms of both geometric and semantic accuracy, when compared to its best match in the ground-truth map, as well as the number of provided cuboids with no matching ground-truth equivalent and vice verse.
The final OMQ score is between 0 and 1 with 1 being the best score.

Current results from the RVSU Semantic SLAM challenge have shown that while the challenge is simple in concept, there is still room for improvement from current state-of-the-art methods.
The highest result for semantic SLAM achieved was 0.39 OMQ when using ground-truth pose data and passive control.
When digging deeper into the results provided, we can see that although the quality of matching cuboids is often good (pairwise quality of up to 0.72) there are too many unmatched cuboids to get a high score.
When competitors bridge the gap from passive to active control, we also commonly see a drop in OMQ of approximately 0.06 despite having more control of the robot's observations.
Those who participated in both passive and active control versions of the semantic SLAM task, focused their research in how to map a scene given a sequence of inputs, rather than how to actively explore to maximize understanding of the scene.
These results suggest that potentially the most fruitful areas for future research lie in better filtering out cuboids that do not match any true object, and in how to best exploit active robot control to improve scene understanding.
There is also yet to be an attempt at solving this challenge using active control and noisy pose estimation which adds further difficulty to the challenge.

\subsubsection{Audio-Visual Navigation}

\begin{figure}[h]
    \centering
    \includegraphics[width=\linewidth]{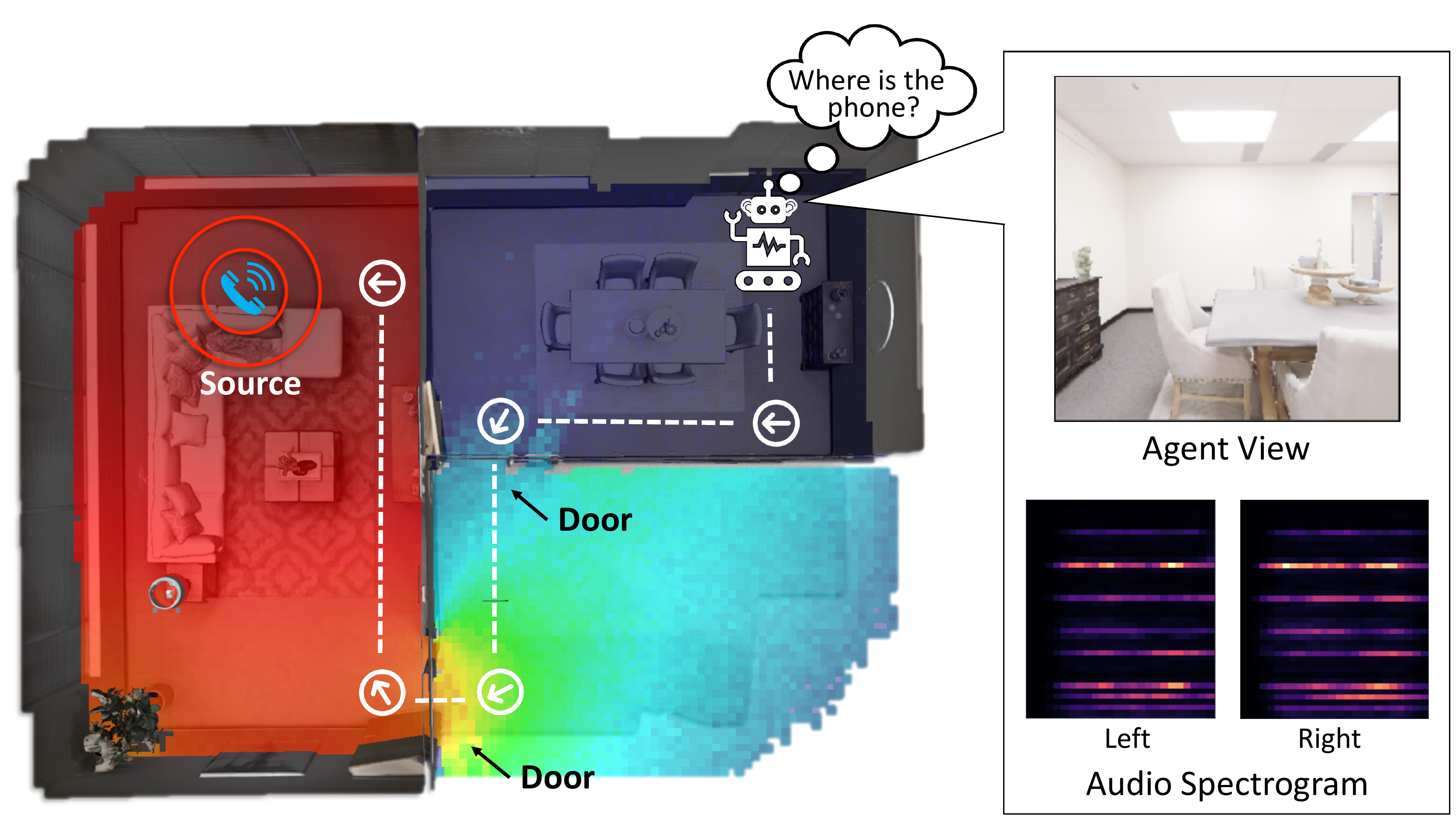}
    \caption{\emph{AudioGoal} tasks an autonomous agent to find an audio source in an unmapped 3D environment by navigating to a goal. Here the top down map is overlaid with the acoustic pressure field heatmap. While audio provides rich directional information about the goal, and audio intensity variation is correlated with the shortest path distance, vision reveals the surrounding geometry in the form of obstacles and free space. An AudioGoal navigation agent should intelligently leverage the synergy of these two complementary signals to successfully navigate in the environment.}
    \label{fig:soundspaces_concept}
\end{figure}

Moving around in the real world is a multi-sensory experience, and an intelligent agent should be able to see, hear and move to successfully interact with its surroundings. While current navigation models tightly integrate seeing and moving, they are deaf to the world around them, motivated by these factors, the audio-visual navigation task was introduced~\cite{gan2019look,chen_soundspaces_2020}, where an embodied agent is tasked to navigate to a sounding object in an unknown unmapped environment with its egocentric visual and audio perception (Figure~\ref{fig:soundspaces_concept}). This audio-visual navigation task can find applications in assistive and mobile robotics, e.g., robots for search and rescue operations and assistive home robots. Along with the task, the SoundSpaces platform was also introduced, a first-of-its-kind audio-visual simulator where an embodied agent could move around in the simulated environment while seeing and hearing.

Audio-visual navigation is a challenging task because the agent not only needs to perceive the surrounding environment, but also to reason about the spatial location of the sound emitter in the environment via the received sound. This new multimodal embodied navigation task has gained attention over the past few years and different methods have been proposed to solve this task, including learning hierarchical policies~\cite{chen_waypoints_2020}, training robust policies with adversarial attack~\cite{YinfengICLR2022saavn} or data augmentation for generalization to novel sounds~\cite{dynamic_av_nav}. However, the performance of SOTA audio-visual navigation models is still not perfect, and thus we organized the SoundSpaces Challenge~\footnote{\url{https://soundspaces.org/challenge}} at CVPR 2021 and 2022, which aims to promote research in the field of developing autonomous embodied agents that are capable of navigating to sounding objects of interest using audio and vision. 

More specifically, in an AudioGoal navigation episode, a sound source is placed at a random location in the environment, and the agent is also positioned with a random pose (location and orientation) at the start of the episode. The agent is tasked to navigate to the sounding object with one of the four actions from the action space: \emph{Move Forward}, \emph{Rotate Left}, \emph{Rotate Right}, and \emph{Done}. At each episode step, the agent receives egocentric (noiseless) RGB-D images captured with a $90^\circ$ field-of-view (FoV) camera, the binaural audio received by the agent. The episode terminates when the agent executes the \emph{Done} action, or it runs out of a pre-specified time budget. The agent is evaluated using standard embodied navigation metrics, such as Success Rate (SR) and SPL~\cite{anderson_arxiv18}. We use SPL as the metric for ranking challenge participants. 

We set up the AudioGoal navigation task on the Matterport3D (MP3D)~\cite{Chang3DV2017Matterport} scene dataset, split into train/val/test splits in 59/10/12 for this challenge due to its large scale. SoundSpaces provides audio renderings for MP3D in the form of pre-rendered room impulse responses (RIRs), which are transfer functions that characterize how sound propagates from one point in space to another point in space. For all MP3D scenes, SoundSpaces discretizes them into grids of spatial resolution 1 meter $\times$ 1 meter and provide RIRs for all pairs of grid points. For the source sound, we use 73/11/18 disjoint sounds in our train/val/test splits, respectively. Each sound clip is 1 second long. The received sound at every step is the result of convolution between the source sound and the RIR corresponding to the source location and current agent pose in the scene. While the \emph{Move Forward} action takes the agent forward by 1 meter in the direction it's currently facing if there is a navigable node in the scene grid in that direction, \emph{Rotate Left} and \emph{Rotate Right} rotate the agent by $90^\circ$ in the clockwise and anti-clockwise directions, respectively. The episode terminates when the agent issues the \emph{Done} action, or it exceeds a budget of 500 steps. 

In SoundSpaces Challenge 2021 and 2022, a total of 25 teams showed interest and 8 teams participated. For SoundSpaces Challenge 2022's leading teams, we observed some similarities between the model design of the top two teams. Both models used a hierarchical navigation architecture (inspired by AV-WaN~\cite{chen_waypoints_2020}), where a high-level (long-term) planner predicts a navigation waypoint in the local neighborhood of the agent at each step, and a low-level (short-term) planner executes atomic actions, such as \emph{Move Forward} and \emph{Rotate Left}, to take the agent to the predicted waypoint. Further, agents that leverage the audio-visual cues from the full $360^\circ$ FoV and train a separate model for stopping are more successful and efficient than the others. Moreover, training an AudioGoal navigation agent in the presence of distractor sound sources also results in learning robust navigation policies that boost navigation performance. The presentation videos from the leading teams can be found on the challenge website.

\begin{figure}[h]
    \centering
    \includegraphics[width=\linewidth]{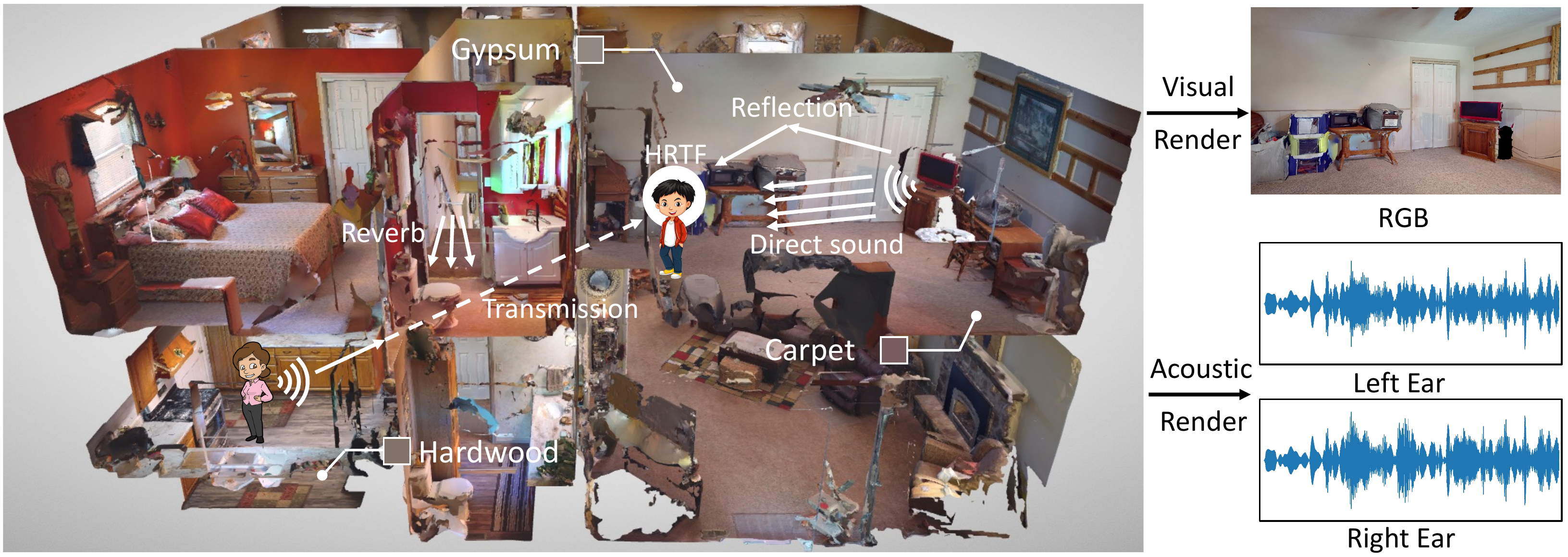}
    \caption{SoundSpaces 2.0, a continuous, configurable, and generalizable audio-visual simulation platform. It models various acoustic phenomena and renders visual and audio observations with spatial and acoustic correspondence.}
    \label{fig:ss2}
\end{figure}

One of the limitations of the SoundSpaces platform is that it provides pre-rendered RIRs for fixed grid points and does not allow users to render sounds for arbitrary locations or environments. To tackle this issue, we have introduced SoundSpaces 2.0~\cite{chen22soundspaces2} (Fig.~\ref{fig:ss2}, a continuous, configurable and generalizable simulator. This new simulator has enabled continuous audio-visual navigation as well as many other embodied audio-visual tasks. We believe this simulator will take the audio-visual navigation task to the next step. Another important direction for future research is for the agent to reason about the semantics between the sound and objects (\eg semantic audio-visual navigation~\cite{chen2021savi} and finding fallen objects~\cite{gan2022finding}). If the agent could leverage the semantics of sounding objects, it could navigate faster by reasoning where the object is located in space based on its category information.

We believe studying audio-visual embodied AI is of vital importance for building truly autonomous robots with rich perception modalities in the real world.

\subsection{Rearrangement Challenges}

This section discusses rearrangement challenges. Rearrangement is described as a canonical task in Embodied AI that may lead to learning representations that are for many downstream tasks~\cite{batra2020rearrangement}. Here, the agents goal is to move or detect the changes from one state of the scene to another. For examples, several objects, such as an apple and a banana may move, and the agent is tasked with detecting that they moved and putting them back to their correct locations.

\noindent
\begin{figure}[h!]
    \centering
    \includegraphics[width=.46\textwidth]{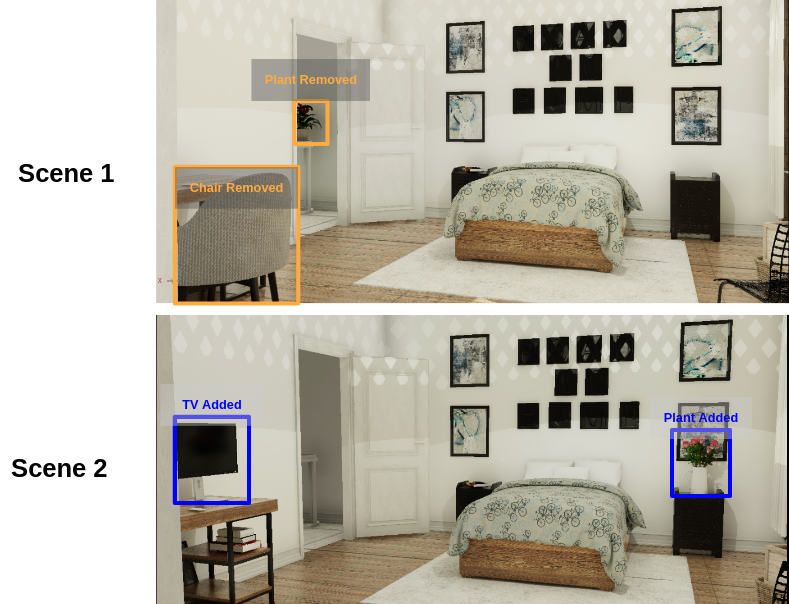}
    \caption{Example of the scene change detection challenge. Between two scenes some objects are added (blue) and removed (orange) and these need to be identified and mapped out.}
\end{figure}

\subsubsection{Scene Change Detection} 

The RVSU scene change detection (SCD) challenge, an extension to the RVSU semantic SLAM challenge, requires identification and mapping of objects which have been added and removed between two traversals of the same base scene~\cite{hall2020robotic}.
Human environments are inherently non-static with objects frequently being added, removed, or shifted.
In order to operate within said environments whilst utilising object maps, it becomes important to be able to identify when these changes have occurred.
This challenge examines perhaps the simplest of such scenarios, where some objects are added or removed while all others remain fixed in place.

The setup for SCD is similar to that shown for the RVSU Semantic SLAM challenge described previously but with some differences in challenge setup within BenchBot.
The SCD challenge also uses the BEAR dataset~\cite{hall2022bear} which already has multiple variants of a set of base scenes.
Variants differ in some objects are added and some removed (as desired for the SCD task), and also there are some lighting variations to increase challenge difficulty.
BenchBot enables the switching between environment variants within one SCD submission as soon as the robot agent determines it is finished with its first traversal.
BenchBot supplies the same robot control that progresses from passive to active control via discrete actions and wherein each action is followed by an observation containing RGB-D images, laser scan, and either ground-truth or estimated robot pose data.
The SCD challenge utilises a variant of the OMQ evaluation measure~\cite{hall2020robotic} which evaluates the final 3D object cuboid map output as part of the challenge.
This variant introduces the necessity for the map to provide an estimate for the likelihood that an object has been added or removed from the scene between traversals.
This state estimation for the object is then combined with the estimation of the label and location of the object to make up the object-level quality score.
As before, the best OMQ score possible is 1 and the worst is 0.

There has been limited engagement with the SCD challenge and there is much room for improvement.
In the CVPR 2022 iteration of this challenge the highest OMQ  score achieved was 0.25.
This is quite lower than the best OMQ score of the semantic SLAM challenge which was able to reach OMQ of 0.39.
This can be attributed somewhat to the approach that competitors used in solving SCD.
All SCD submissions performed semantic SLAM on the two different traversals and did a naive comparison of the resultant cuboid maps.
This led to an accumulation of the errors seen across the maps for both traversals.
This simple beginning shows that there are many directions that can still be experimented with in order to improve SCD in future years.
This may include more targeted approaches to navigation and/or mapping within the second traversal which utilises the scene knowledge from the first traversal.
There is still much research to be done in how to reliably identify and map out changes between scenes.

\noindent
\begin{figure}[ht!]
    \vspace{0.1in}
    \centering
    \includegraphics[width=0.47\textwidth]{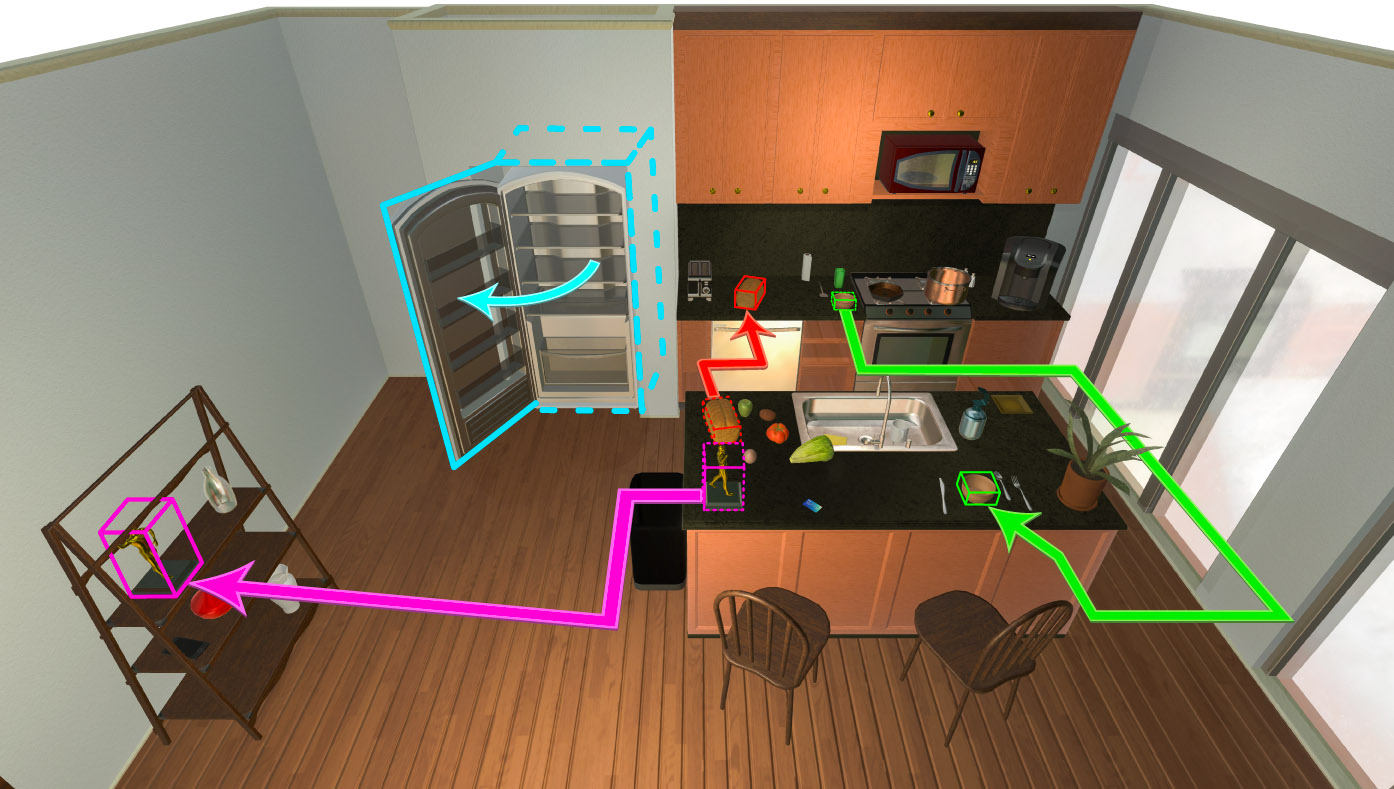}
    \vspace{-0.1in}
    \captionsetup{type=figure}
    \caption{\textbf{AI2-THOR Visual Room Rearrangement Challenge.} An agent must change pose and attributes of objects in a household environment to restore the environment to an initial state.}
    \vspace{-0.1in}\label{fig:rearrangement}
\end{figure}

\subsubsection{Interactive Rearrangement}

While the PointNav and ObjectNav tasks have led to substantial advances in embodied AI, performance on these tasks has steadily improved with PointNav being nearly solved~ \cite{ramakrishnan_arxiv21}. In light of this fast progress, researchers from nine institutions proposed the \emph{rearragement} as the next frontier for research in embodied AI~ \cite{batra2020rearrangement}. At a high level, in rearrangement, an embodied agent must interact with its environment to transform the environment from it's initial state $s^{\text{init}}$ to a goal state $s^{\text{goal}}$. This general formulation of rearrangement leaves much unspecified, namely: (1) which environment? (2) what affectors/actions are available to agent? (3) how are states $s^{\text{init}},s^{\text{goal}}$ specified? Given the EAI community's focus on building agents capable of assisting humans in everyday tasks, all existing instantiations of the rearrangement task embody agents in household environments and focus on object-based rearrangement: the difference between goal and initial environment states is confined to objects pose (position/rotation) and attributes (e.g. is the object opened or closed?). Successful rearrangement in these tasks requires agents to flexibly encode environment environment states, to dynamically update these encodings as they interact with their environment, and also to making long-term plans (frequently of the traveling-salesman variety) to maximize the efficiency of rearrangement. We now detail the two rearrangement challenges, AI2-THOR Visual Room Rearrangement and TDW-Transport, held at the EAI workshop in past years.

The AI2-THOR Visual Room Rearrangement (RoomR) task~\cite{weihs2021rearrangement} occurs in two phases, see Figure~\ref{fig:rearrangement}. In the \emph{Walkthrough phase} the the agent explores a room and builds an internal representation of the room's configuration ($s^{\text{goal}}$). Then, in the \emph{Unshuffle phase}, the agent is placed within the same environment but objects within this environment have been randomly moved to different locations and opened/closed ($s^{\text{init}}$), the agent must now restore objects back to their original states. As this 2-phase RoomR is quite challenging, a 1-phase variant was also proposed where the agent enacts the Walkthrough and Unshuffle phases simultaneously, receiving egocentric RGB-D images of the environment in both the $s^{\text{init}}$ and $s^{\text{goal}}$ states at each step. In the 2021 RoomR challenge, no participants were able to outperform the baseline model, which used a 2D semantic mapping approach along with imitation learning from a heuristic expert agent. In 2022 however, several exciting approaches were released resulting in dramatic improvements in performance. For the 1-phase variant, performance leapt from ${\approx}9\%$ to ${\approx}24\%$ on the \textsc{FixedStrict} metric on the test-set. Advances making this possible included (1) the use of CLIP-pretrained visual encoders~\cite{khandelwalEtAl2021embodiedclip} and (2) large-scale pre-training using procedurally generated environments~\cite{deitke2022procthor}. Unlike the end-to-end approaches used for the 1-phase variant, the most successful methods for the 2-phase variant used powerful inductive biases in the form of semantic mapping and planning algorithms. In a yet unpublished work, 2022 2-phase challenge winner used voxel-based 3D semantic map and shortest path planners to bulid an agent attaining ${\approx}15\%$ \textsc{FixedStrict} on the test-set (dramatically beating the baseline performance of $<1\%$). The differences between the approaches used in the 1- and 2-phase variants is striking: it seems that new algorithms are required to bring fully end-to-end methods to the challenging 2-phase setting.

\noindent
\begin{figure}[ht!]
    \vspace{0.1in}
    \centering
    \includegraphics[width=0.47\textwidth]{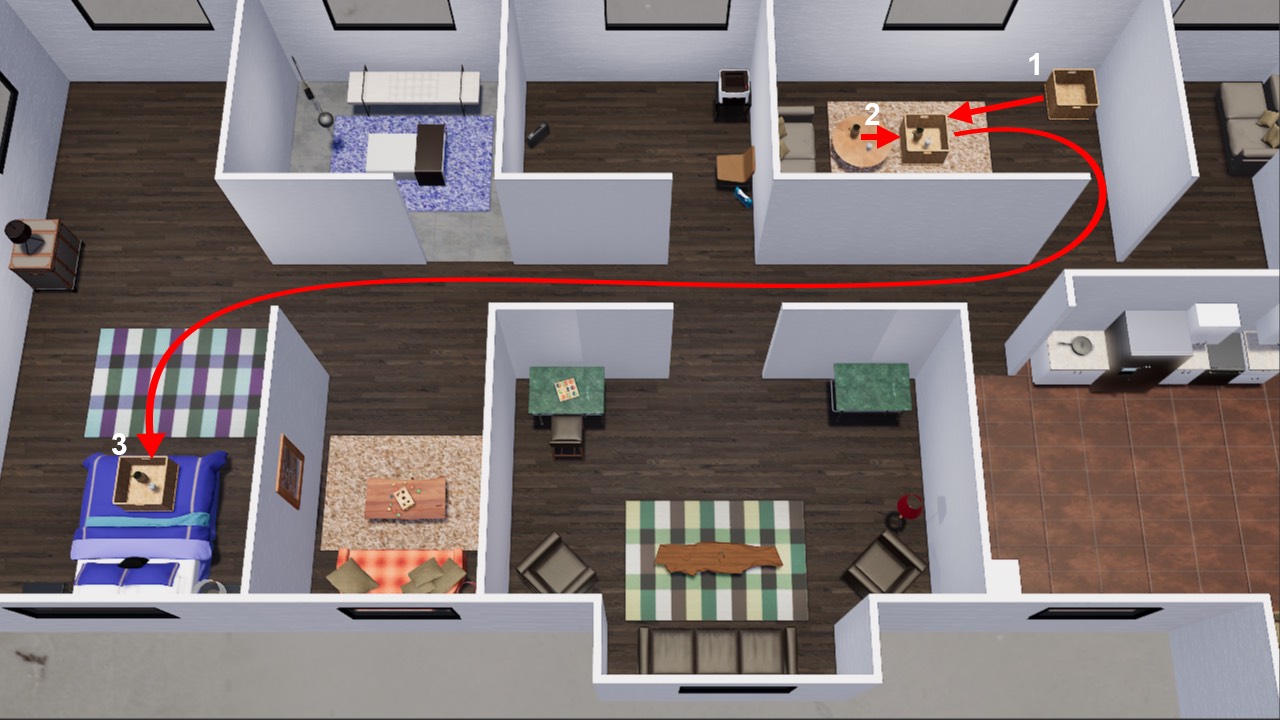}
    \vspace{-0.1in}
    \captionsetup{type=figure}
    \caption{\textbf{TDW-Transport Challenge.} In this example task, the agent must transport two objects on the table in one room and place them on the bed in the bedroom. The agent can first pick up the container, put two objects into it, and then transport them to the target location.}
    \vspace{-0.1in}\label{fig:TDW-transport}
\end{figure}

TDW-Transport Challenge~\cite{gan2022threedworld} is an object-goal driven interactive navigation task (see Figure ~\ref{fig:TDW-transport}). In this challenge, an embodied agent is spawned randomly in a house and is required to find a small set of objects scattered around the house and transport them to a desired final location.
We also position various containers around the house; the agent can find these containers and place some objects into them. Without using a container as a tool, the agent can only transport up to two objects at a time. However, using a container, the agent can collect several objects and then transport them together. While the containers help the agent transport more than two items, it also takes some time to find them. Therefore, the agent has to decide to use containers or not.

The embodied agent is equipped with an RGB-D camera. There are two types of actions of the agent: navigation and interactive actions. Navigation actions include Move Forward($\alpha$ meters), Turn Left($\theta$ degrees), Turn Right($\theta$ degrees). Interactive actions include Reach to object, Put into container, Grasp, and Drop.  The objective of this challenge is to transport the maximum number of objects in fixed steps as efficiently as possible. We use the transport rate as an evaluation metric, which measure the fraction of the objects successfully transported to the desired position within a given budget.

\subsection{Embodied Vision-and-Language}

This section discusses the embodied vision-and-language challenges. In each challenge, natural language is used to convey the goal to the agent. For example, the agent may be tasked with following instructions to complete a task. Since language is the primary means of human communication, advances in embodied vision-and-language research will make it easier for a human to naturally interact with the trained agents. Additionally, language imposes a data-sparse regime, as examples cannot be created automatically, as precision in language is directly tied to a specific scene layout (e.g. ``on the left/right"), and it is an open challenge as to if unimodal representations can be leveraged in this embodied space \cite{bisk2020}.

\noindent
\begin{minipage}[l]{0.46\textwidth}
    \vspace{0.1in}
    \centering
    \includegraphics[width=\textwidth]{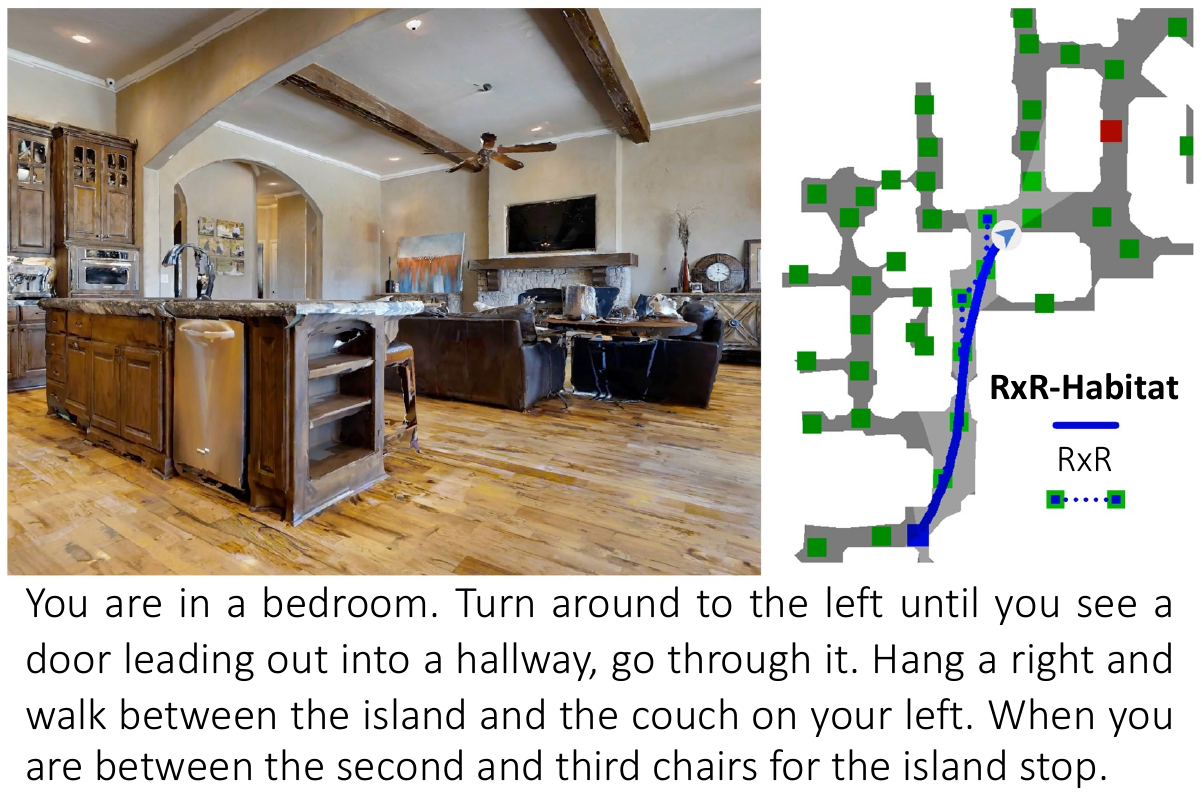}
    \vspace{-0.1in}
    \captionsetup{type=figure}
    \captionof{figure}{The Room-Across-Room Habitat Challenge (RxR-Habitat) is a multilingual instruction-following task set in simulated indoor environments requiring realistic navigation over long action sequences.}
    \vspace{-0.1in}
    \label{fig:rxr_habitat_challenge}
\end{minipage}

\subsubsection{Navigation Instruction Following}
Navigation guided by natural language has long been a desired foundational ability of intelligent agents. In Vision-and-Language Navigation (VLN), an agent is given egocentric vision in a realistic, previously-unseen environment and tasked with following a path described in natural language, \eg, \textit{Move toward the dining table. Go down the hallway toward the kitchen and stop at the sink}. The Room-Across-Room Habitat Challenge (RxR-Habitat) instantiates VLN in simulated indoor environments, provides multilingual instructions, and requires agents to navigate via long action sequences in a realistic, continuous 3D world (Figure~\ref{fig:rxr_habitat_challenge}). Solving RxR-Habitat would have applications in many domains, such as personal robotic assistants, and lead to a better scientific understanding of the connection between language, vision, and action.

The RxR-Habitat Challenge takes place in 3D reconstructions of Matterport3D scenes \cite{matterport3d} and interacts with those scenes using the Habitat Simulator \cite{savva2019habitat}. We model the agent embodiment after a robot of radius 0.18m and height 0.88m with a camera mount at 0.88m. An episode is specified by a scene, a start location, a language instruction, and the implied path.
At each time step, the agent observes egocentric vision in the form of a single forward-facing, noiseless 480x640 RGB-D image with a 79$^{\circ}$ HFOV. The agent also receives the natural language instruction from one of three languages: English, Hindi, or Telugu. The action space is discrete and noiseless, consisting of actions $\{$\texttt{MOVE\_FORWARD}, \texttt{TURN\_LEFT}, \texttt{TURN\_RIGHT}, \texttt{STOP}, \texttt{LOOK\_UP}, \texttt{LOOK\_DOWN}$\}$. Forward movement is 0.25m and turning and looking actions are performed in 30$^{\circ}$ increments. Actions that result in collision terminate upon collision, \ie, no wall sliding. An episode ends when the agent calls \texttt{STOP}.

The dataset used in RxR-Habitat is the Room-Across-Room (RxR) dataset \cite{rxr} ported from high-level discrete VLN environments \cite{anderson_cvpr18} to the continuous VLN-CE environments \cite{krantz_vlnce_2020} used in Habitat. The dataset is split into training (Train: 60,300 episodes, 59 scenes), validation in environments seen during training (Val-Seen: 6,746 episodes, 57 scenes), validation in environments not seen during training (Val-Unseen: 11,006 episodes, 11 scenes), and testing in environments not seen during training (Test-Challenge: 9,557 episodes, 17 scenes), each with a roughly equal distribution between English, Hindi, and Telugu instructions. To submit to the RxR-Habitat leaderboard~\footnote{\url{https://ai.google.com/research/rxr/habitat}}, participants run inference on the Test-Challenge split and submit the inferred agent paths. The leaderboard evaluates these paths against held-out ground-truth paths. Agent performance is reported as the average of episodic performance. The official comparison metric between the agent's path and the ground truth path is normalized dynamic time warping (nDTW) \cite{magalhaes2019effective} which scores path alignment between 0 and 1 with 1 indicating identical paths. Additional metrics reported for analysis include path length (PL), navigation error (NE), success rate (SR) and success weighted by inverse path length (SPL)~\cite{anderson_arxiv18}.

RxR-Habitat is incredibly difficult; the interplay between perception, control, and language understanding makes instruction-following an interdisciplinary problem. Realistic environments and unconstrained natural language lead to a long tail of vision and language grounding, and the low-level action space makes learning the relationship between instructions and actions highly implicit. The RxR-Habitat Challenge took place in 2021 and again in 2022. The baseline model is a cross-modal attention (CMA) model \cite{krantz_vlnce_2020} that attends between vision and language encodings, predicts actions end-to-end from observation, and is trained with behavior cloning (nDTW: 0.3086). In the first year, teams failed to surpass the performance of this baseline. However, a significant improvement in SOTA was attained in 2022; the top submission (Reborn \cite{an20221st}) produced an nDTW of 0.5543 --- an 80\% relative improvement over the baseline. This was enabled by an effective hierarchy of waypoint candidate prediction, waypoint selection (the discrete VLN task), and waypoint navigation. For waypoint selection, a history-aware transformer was trained in discrete VLN with augmentations including synthetic instructions, environment editing, and ensembling. It was then transferred and tuned in continuous environments. Despite this remarkable improvement, a performance gap still exists between SOTA in continuous versus discrete environments, with human performance even higher. Evidently, this direction of research is still far from saturated.

\noindent
\begin{minipage}[l]{0.46\textwidth}
    \vspace{0.1in}
    \centering
    \includegraphics[width=\textwidth]{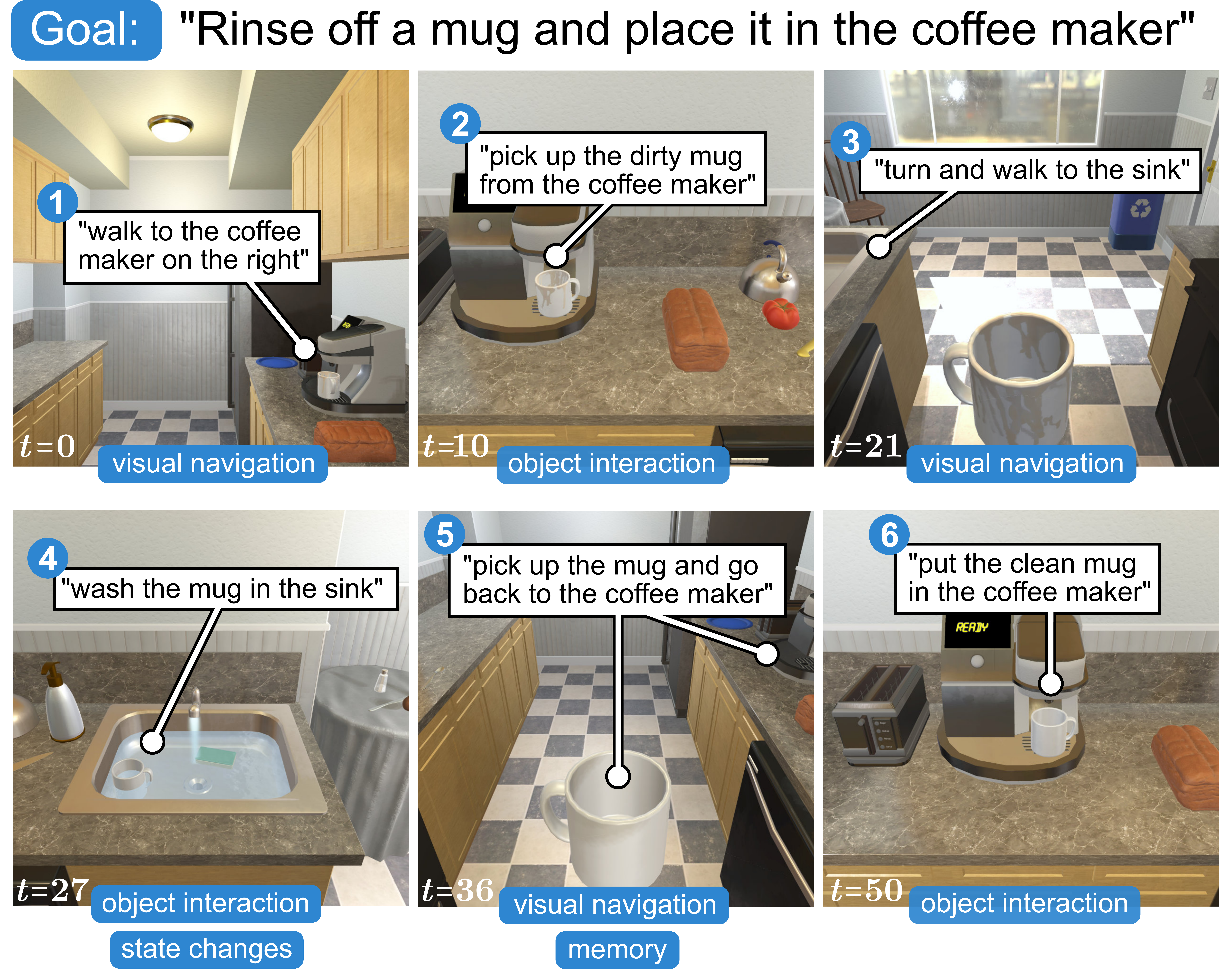}
    \vspace{-0.1in}
    \captionsetup{type=figure}
    \captionof{figure}{ALFRED involves interactions with objects, keeping track of state changes, and references to previous instructions. The dataset consists of 25k language directives corresponding to expert demonstrations of household tasks.
    We highlight several frames corresponding to portions of the accompanying language instruction.}
    \vspace{-0.1in}
\end{minipage}

\subsubsection{Interactive Instruction Following.} 

ALFRED is a benchmark for connecting human language to \textit{actions}, \textit{behaviors}, and \textit{objects} in interactive visual environments.
Planner-based expert demonstrations are accompanied by both high- and low-level human language instructions in 120 indoor scenes in AI2-THOR. 
These demonstrations involve partial observability, long action horizons, underspecified natural language, and irreversible actions. 

The dataset includes over 25K English language directives describing 8K expert demonstrations averaging 50 steps each, resulting in >428K image-action pairs.
Motivated by work in robotics on segmentation-based grasping, agents in ALFRED interact with objects visually, specifying a pixelwise interaction mask of the target object.
This inference is more realistic than simple object class prediction, where localization is treated as a solved problem.
Existing beam-search and backtracking solutions are infeasible due to the larger action and state spaces, long horizon, and inability to undo certain actions. Agents are evaluated on their ability to achieve directives in both seen and unseen rooms. Evaluation metrics include: success rate (SR), success weighted by path-length (SPL), and Goal-Condition success which measures completed subtasks.

Current state-of-the-art approaches in ALFRED use spatial-semantic mapping  \cite{blukis2022persistent,min2021film} to explore and build persistent representations of the environment before grounding instructions. These representations have also been coupled with symbolic planners and modular policies for better generalization to unseen rooms. Currently, the best performing agent achieves 40\% success in seen rooms and 36\% in unseen rooms.

\noindent
\begin{figure}[h!]
    \centering
    \includegraphics[width=.45\textwidth]{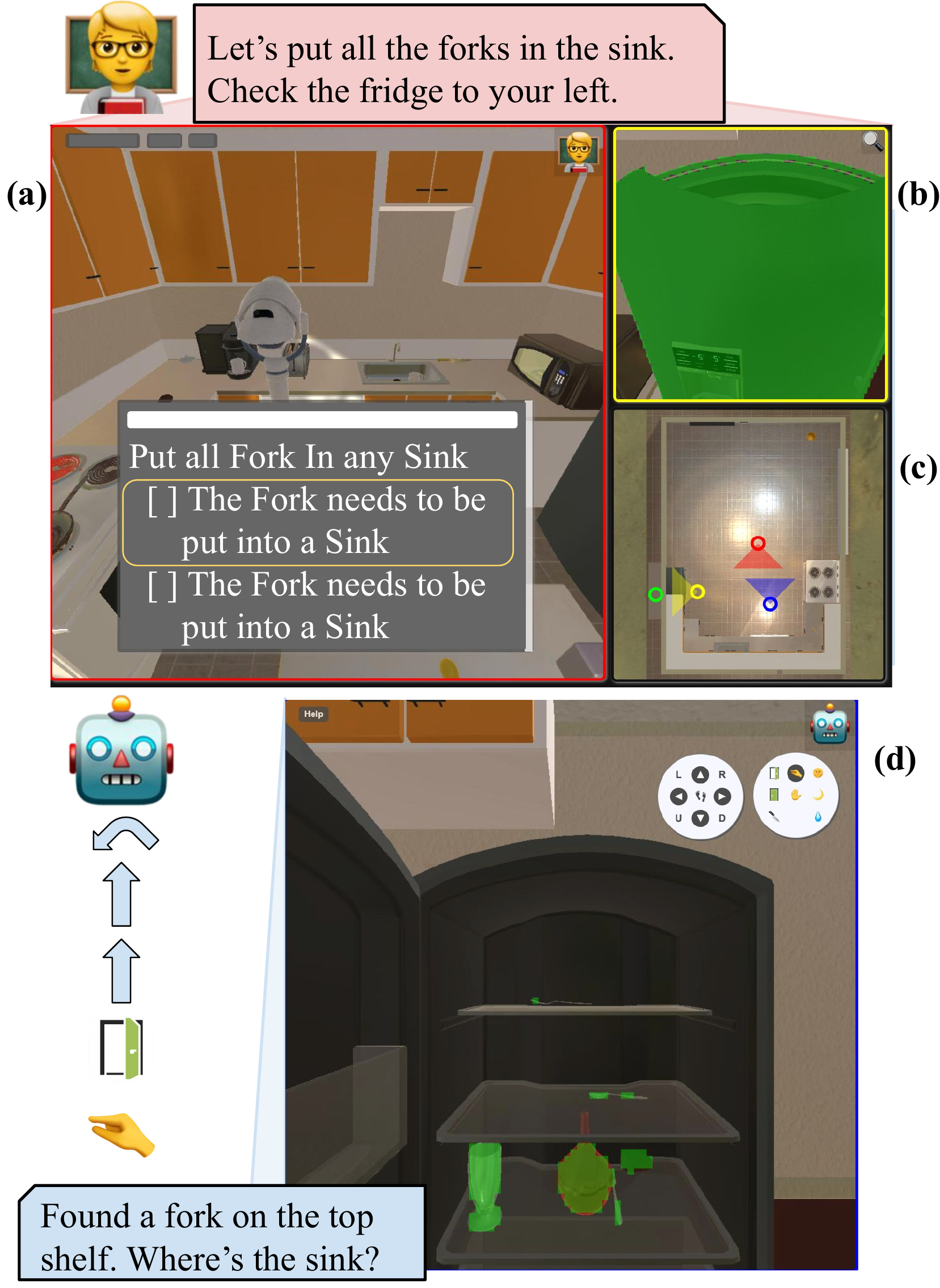}
    \caption{
        In the TEACh Two-Agent Task Completion challenge, the \teachcommander\ has oracle task details (a), object locations (b), a map (c), and egocentric views from both agents, but cannot act in the environment, only communicate.
        The \teachfollower\ carries out the task and asks questions (d).
        The agents can only communicate via text.}
    \label{fig:teach_teaser}
\end{figure}

\subsubsection{Interactive Instruction Following with Dialog}
\textbf{T}ask-driven \textbf{E}mbodied \textbf{A}gents that \textbf{C}hat (TEACh) is a dataset of over 3,000 human--human, interactive dialogues and demonstrations of household task completion in the AI2-THOR simulator.
Robots operating in human spaces must be able to engage in such natural language interaction with people, both understanding and executing instructions and using conversation \cite{thomason:corl19,Roman2020} to resolve ambiguity \cite{Nguyen2022} and recover from mistakes.
A \teachcommander\ with access to oracle information about a task communicates in natural language with a \teachfollower. 
The \teachfollower\ navigates through and interacts with the environment to complete tasks varying in complexity from \texttt{Make Coffee} to \texttt{Prepare Breakfast}, asking questions and getting additional information from the \teachcommander\ (Figure~\ref{fig:teach_teaser}).

There are 12 task types in TEACh with 438 unique combinations of task parameters (e.g., \texttt{Make Salad} with 1 versus 2 slices of \texttt{Tomato}) in 109 AI2-THOR environments.
On average, there are more than 13 utterances in each cooperative dialogue, with tasks taking an average of 131 \teachfollower\ actions to complete compared to ALFRED's 50 due to both task complexity and non-optimal planning.
A major difference between the TEACh and ALFRED challenge is edge cases in the environments due to ALFRED's rejection sampling: if a PDDL planner could not resolve an ALFRED task given an initial scene configuration, it was rejected from data, where TEACh scene configurations are rejected only when a \textit{human} cannot resolve them.
This decision results in many ``corner cases'' in TEACh that require human ingenuity, for example filling a pot with water using a cup as an intermediate vessel when the pot itself is too large to fit in the sink basin.

The Two-Agent Task Completion (TATC) challenge is based on the TEACh data, and involves modeling \textit{both} the \teachcommander\ and \teachfollower\ agents, which have distinct action and observation spaces but a common household task goal.
The \teachcommander\ agent has access to a structured representation of the goal and its component parts, as well as search functions to identify the locations and physical appearance of objects in the environment by class or id.
The \teachfollower\ is analogous to an ALFRED agent, but with a wider action space that includes, for example, pouring liquids from one container to another.
Further, object interactions are done via individual $(x,y)$ coordinate predictions, rather than the full object masks used in ALFRED, analogous to the click inputs of human users who provided demonstrations.
The agents both have a \texttt{communicate} action that adds to a mutually-visible dialogue history, and requires generating text.

TATC agents are evaluated via SR and SPL, similar to ALFRED agents.
Rule-based, planning agents for TATC achieve about 24\% SR, with planning corner cases dominating failures.
A learned \teachfollower\ based on the Episodic Transformer~\cite{pashevich:et} with a rule-based, simple \teachcommander\ that simply reports the raw text of the next task subgoal as a communication action achieves nearly 0\%.
We are eager to see whether mapping-based approaches like those succeeding at ALFRED can adapt to the wider space of tasks and environment corner cases in TEACh.

\section{Common Approaches}

This section presents common approaches used by the winners of the challenges. We discuss large-scale training by scaling up datasets and compute, leveraging visual pre-trained models such as CLIP, the use of inductive biases such as maps, goal embeddings to represent different tasks, and visual and dynamic augmentation to make simulators more noisy and closer to reality.

\subsection{Large-Scale Training}%
Embodied AI is seeing the same trend as computer vision and natural language processing, where massive datasets and more computing power enable higher-performing models.

Massive datasets have been obtained by ProcTHOR, which trains on 10K procedurally generated houses~\cite{deitke2022procthor}; HM3D, which captures 1K static scans of real-world environments~\cite{ramakrishnan2021habitat}; and Habitat-Web, which builds an Amazon Mechanical Turk task to collect 80K imitation learning examples of humans performing ObjectNav in simulated environments~\cite{ramrakhya2022habitat}. ProcTHOR supports object interaction, and training a simple RGB model with it using on-policy RL led to state-of-the-art results for the Habitat 2022 ObjectNav Challenge, the RoboTHOR ObjectNav challenge, and the AI2-THOR Rearrangement Challenge. Moreover, 0-Shot performance, with models pre-trained on ProcTHOR, often beats the same models trained on the training data from the benchmark it is evaluated on. The scale and diversity of HM3D led to models trained on it achieving state-of-the-art performance for PointNav models when evaluated on Gibson~\cite{xiazamirhe2018gibsonenv}, MP3D~\cite{chang2017matterport3d}, and HM3D~\cite{ramakrishnan2021habitat}. Using imitation learning to train on Habitat-Web led to state-of-the-art results in the 2021 Habitat ObjectNav Challenge, which later improved its performance with online fine-tuning. We expect the trend of building and training on massively larger datasets to continue leading to better generalization.

Simultaneously, many of the top approaches to these challenges are scaling compute to train on hundreds of millions to billions of steps. DD-PPO presented an on-policy RL method that has been used throughout embodied AI to train agents in a distributed manner~\cite{wijmans2019dd, weihs2020allenact}. It scaled PointNav to train for billions of steps across 64 GPUs and showed near-perfect performance in PointNav in unseen environments with just an RGB-D camera and a GPS+Compass sensor. For ObjectNav training, ProcTHOR similarly trained with DD-PPO for 420 million steps, and was later fine-tuned for 195M steps on HM3D~\cite{ramakrishnan2021habitat} and fine-tuned for 29 million steps on RoboTHOR~\cite{deitke2020robothor}. Habitat-Web used behavior cloning to train for 400 million steps for their 2021 Habitat Challenge entry. With the growing size of datasets, and the added performance gained by training for orders of magnitude longer, we suspect further scaling compute to lead to better performing agents.

\subsection{Visual Pre-Training}
Initial successes in deep reinforcement learning were largely focused on graphically simplistic environments, \eg Atari games, for which complex visual processing was, in large part, unnecessary. For instance, the seminal work of Mnih et al.~\cite{mnih_nature15} achieved human-level performance on dozens of Atari games using used a model with only three convolutional layers. Several initial works in embodied AI, in part due to computational constraints when training RL agents, adopted this mindset; for instance, Savva et al.~\cite{habitat19iccv} trained models using a 3-layer image processing CNN for Point Navigation. As embodied agents are, ostensibly, meant to be embodied in the real-world, one might expected the they would benefit from image processing architectures designed for use with real images and, indeed, this has proven to be the case. A recent work has shown that modifying existing embodied baseline models by replacing their visual backbones with a CLIP-pretrained ResNet-50 can result in dramatic improvements~\cite{khandelwalEtAl2021embodiedclip}. The top performing models of 1-Phase Rearrangement, RoboTHOR ObjectNav leaderboards, and Habitat ObjectNav leaderboard, use variants of this ``EmbCLIP'' architecture~\cite{deitke2022procthor}. Several other top performing models to other challenges use pretrained vision models for object detection and semantic segmentation (RVSU Semantic SLAM, MultiON, and Two-Phase Rearrangement).

\setlength{\tabcolsep}{4pt}
\begin{table*}[]
\small   
\begin{tabular}{lllllclllc}
\hline
\textbf{}                         & \textbf{}          & \textbf{} & \multicolumn{3}{c}{\textbf{Best End-to-end}}       & \textbf{} & \multicolumn{3}{c}{\textbf{Best Modular}}          \\
\textbf{Challenge}                & \textbf{Simulator} & \textbf{} & \textbf{Method} & \textbf{Success} & \textbf{Rank} & \textbf{} & \textbf{Method} & \textbf{Success} & \textbf{Rank} \\ \hline
ObjectNav                         & Habitat            &           & Habitat-Web     & 60               & 2             &           & Stretch         & 60               & 1             \\
Audio-Visual Navigation           & SoundSpaces        &           & Freiburg Sound  & 73               & 2             &           & colab\_buaa     & 78               & 1             \\
Multi-ON                          & Habitat            &           & -               & -                & -             &           & exp\_map        & 39               & 1             \\
Navigation Instruction Following  & VLN-RxR            &           & CMA Baseline    & 13.93            & 10            &           & Reborn          & 45.82            & 1             \\
Interactive Instruction Following & AI2-THOR           &           & APM             & 15.43            & 14            &           & EPA             & 36.07            & 1             \\
Rearrangement                     & AI2-THOR           &           & ResNet18 + ANM  & 0.5              & 6             &           & TIDEE           & 28.94            & 1             \\ \hline
\end{tabular}
\vspace{-6pt}
\caption{Table summarizing the performance of best end-to-end and best modular methods across various challenges.}
\label{tab:e2e_modular}
\end{table*}

\subsection{End-to-end vs Modular} 
In the last few years, two classes of methods have emerged for various embodied AI tasks: (1) end-to-end and (2) modular. The end-to-end methods learn to predict low-level actions directly from input observations. They typically use a deep neural network consisting of a visual encoder followed by a recurrent layer for memory and are trained using imitation learning or reinforcement learning. Earliest application of end-to-end methods on embodied AI tasks include \cite{ lample2016playing, zhu2017target, mirowski2016learning, chaplot2017arnold, savva2017minos, hermann2017grounded, chaplot2017gated}. End-to-end RL methods have also been scaled to train with billions of samples using distributed training~\cite{wijmans2019decentralized} or using tens of thousands of procedurally generated scenes~\cite{deitke2022procthor}. Researchers have also introduced some structure in end-to-end policies such using spatial representations~\cite{gupta2017cognitive, parisotto2017neural, chaplot2018active, henriques2018mapnet, gordon2018iqa} and topological representations~\cite{yang2018visual, savinov2018semi, savinov2018episodic}. 

Modular methods use multiple modules to break down the embodied AI tasks. Each module is trained for a specific subtask using direct supervision. The modular decomposition typically includes separate modules for perception (mapping, pose estimation, SLAM), encoding goals, global waypoint selection policies, planning and local obstacle avoidance policies. Rather than training all modules end-to-end, each module is trained separately using direct supervision, which also allows use of non-differentiable classical modules within the embodied AI pipeline. Earliest learning-based modular methods include ~\cite{chaplot2020learning, chaplot2020neural, chaplot2020object} which show their effectiveness on various navigation tasks such as Exploration, ImageNav and ObjectNav. Variants of these methods include improvements in mapping by anticipating unseen parts~\cite{ramakrishnan2020occant} or by using density-based maps~\cite{bigazzi2022focus}; and learning global waypoint selection policies in ObjectNav and ImageNav entirely using offline or passive datasets to improve sample and compute efficiency~\cite{ramakrishnan2022poni,hahn_nrns_2021,mezghani2021memory,wassermanlast}. Recently, modular methods have also been applied to longer horizon tasks such as Navigation Instruction Following in VLN-CE~\cite{Krantz_2021_ICCV,an20221st,raychaudhuri2021language}, Interactive Instruction Following in ALFRED~\cite{min2021film, liu9planning, murray2022following}, Rearrangement in AI2 Thor~\cite{sarch2022tidee, trabucco2022simple}, and Rearrangement in Habitat~\cite{kant2022housekeep}.

In Table~\ref{tab:e2e_modular}, we show the performance of best end-to-end and modular methods in various 2022 Embodied AI challenges. The table shows that while end-to-end method performance is comparable to modular methods on easier and relatively shorter horizon tasks such as ObjectNav and Audio-Visual Navigation, the performance gap increases as the complexity of the task increases such as in Interactive Navigation and Rearrangement. This is likely because as the task horizon increases, the exploration complexity increases exponentially when training end-to-end with just reinforcement learning.

\subsection{Visual and Dynamic Augmentation}
Visual and dynamic augmentation of real-world datasets has proven to be a key technique for enabling robotic systems trained in simulation to transfer to unseen environments and even to reality. For years in the robotics and learning community, a prevalent attitude has been that simulation transfers poorly to reality. One justification for this perspective is that the dynamics models of most simulations are not good enough to reveal problems that typically occur in real robotic deployments, such as wheel slippage, odometry drift, floor irregularities, nonlinear motor and dynamic responses, and component breakage and burnout. Another justification is that simulated evaluation can reveal problems with systems, but cannot validate them: validation tests for robotic systems must ultimately be performed on-robot.

Nevertheless, many existing systems have shown successful transfer to novel and to real-world environments by augmenting training datasets with noise, static obstacles, dynamic obstacles, and changes to visual appearance. 
Many approaches add noise to sensors, actions and even environment dynamics, effectively making each episode occur in a  distinctive environment; these techniques have proved useful for translating LiDAR-based policies trained in simulation to the real world \cite{faust2018prm,francis2020long} and for estimating the safety of plans prior to deployment \cite{xiao2021toward}. %
Other approaches improve performance by adding static obstacles to the environment in simulation, also effectively increasing the space of environments trained on \cite{xiao2021toward}.
An interesting example of this presented at the workshop involves training in a simulated environment with variable dynamics and using an adaptation module to perform system identification in real environments \cite{kumar2021rma}, \cite{fu2022coupling}.

However, visual policies present other difficulties: a policy trained on one set of objects and lighting conditions is unlikely to transfer to other objects and conditions~\cite{deitke2022procthor}. Adding noise has been used to improve robustness \cite{fang2019scene}, and the RSVU challenges add distractor objects to reduce the effects of distractors~\cite{hall2020robotic}. The RL-CycleGan approach uses style transfer to make simulated environments appear more like the real world \cite{rao2020rl}. Most recently, ProcTHOR~\cite{deitke2022procthor} attempts to address the visual diversity issue by generating large numbers of synthetic environments.

Finally, while the pandemic disrupted many plans for real-world  deployments, both the iGibson, RoboTHOR, and Habitat challenges included tests of simulation-trained policies in real deployments~\cite{xia2020interactive,deitke2020robothor,batra2020objectnav}. These environments proved challenging for many policies; nevertheless, many policies were still able to function, and going forward tests in the real will be an important validation step for embodied AI agents. As datasets collected from real evaluations increase, the opportunity exists to train policies directly over this real-world data, which has already proved useful in a grasping and manipulation context~\cite{bahl2022human} and for legged locomotion~\cite{smith2022walk}. 

\section{Future Directions}

In this section, we discuss promising future directions for embodied AI, including further leveraging pre-trained models, world models and inverse graphics, simulation and dataset advances, sim2real approaches, procedural generation, generalist agents, and multi-agent and human interaction.

\subsection{Pre-training}

Pre-training has powered impressive results from visual recognition~\cite{girshick2014rich}, natural language~\cite{radford2019language,devlin2018bert}, and audio~\cite{oord2016wavenet}.
Pre-trained models can be repurposed through fine-tuning, zero-shot generalization, or prompting to perform diverse tasks.
However, pre-training has not yet found such levels of success in embodied AI. 
Recent work has begun to explore this direction, showing that pre-trained models can help improve performance, efficiency and expand the scope of solvable tasks.
This section discusses how pre-training can help embodied AI with visual pre-training objectives, the role of scale in pre-training, pre-training for task specification, and pre-trained behavioral priors.

One promising area is new pre-training objectives for visual representations in embodied AI.
Prior work shows supervised pre-training is effective for navigation and manipulation tasks \cite{yen2020learning, shah2021rrl, sax2018mid}.
However, a large-scale study \cite{wijmans2019dd} showed that at scale, supervised pre-training visual representations from ImageNet could hurt downstream performance in PointNav.
EmbCLIP shows that unsupervised pre-training with a pre-trained CLIP visual encoder is effective for various embodied AI tasks \cite{khandelwalEtAl2021embodiedclip}.
Other works explore pre-training with masked auto-encoders~\cite{xiao2022masked}, contrastive learning~\cite{du2021curious,nair2022r3m,sermanet2018time}, or other SSL objectives~\cite{yadav2022offline}.
Future work may explore tailoring pre-training objectives specifically for control.
For example, pre-training may account for the temporal aspect of decision making~\cite{gregor2018temporal}, be embodiment agnostic~\cite{stadie2017third},  curiosity-driven~\cite{du2021curious},or avoid pixel reconstruction~\cite{zhang2020learning}. Analogous to pretrained visual representation for visual navigation, audio-visual representations~\cite{alwassel2020self,Morgado2021AudioVisualID,mittal2022learning} can be adopted for tasks with multi-modal inputs~\cite{gan2019look,chen_soundspaces_2020} in future work.

Another way pre-training may benefit embodied AI is with scaling model and dataset size.
Currently, works use a variety of datasets for pre-training such as Epic Kitchens \cite{damen2022rescaling,damen2018scaling,VISOR2022}, YouTube 100 days of hands \cite{shan2020understanding}, Something-Something \cite{goyal2017something}, Ego4D \cite{grauman2022ego4d}, and RealEstate10k \cite{46965} datasets.
The curation of data for pre-training matters, with pre-training on unlabeled curated datasets outperforming labeled datasets on downstream tasks \cite{xiao2022masked}.
Increasing model size also promises benefits, with larger ResNet showing better performance \cite{wijmans2020train}.
Prior work pre-trains ResNet-50 \cite{nair2022r3m,khandelwalEtAl2021embodiedclip,yadav2022offline}, CLIP \cite{khandelwalEtAl2021embodiedclip}, or ViT models \cite{xiao2022masked}.
With the success of neural scaling laws~\cite{kaplan2020scaling} in vision and language, future work in embodied AI may translate these lessons to pre-training larger models with larger datasets.

Pre-training also provides a way to specify diverse tasks for agents easily.
Open-world agents must be able to flexibly complete tasks with unseen goals or task specifications.
Prior work shows that pre-trained models can provide dense reward supervision \cite{cui2022can, shao2021concept2robot, chen2021learning}. 
Other work shows that pre-trained models can be leveraged for open-world object detection, allowing for zero-shot generalization to new goals in navigation tasks \cite{al2022zero,gadre2022clip,majumdar2022zson}. 
Finally, some methods explore generalization to new language instructions by employing pre-trained models \cite{shridhar2022cliport}. 
There are further opportunities to use such models for zero-shot generalization to completing new tasks, new goals, or flexibly specifying goals in different input modalities.

Finally, pre-training can learn behavioral priors for interaction.
The previously discussed pre-training objectives primarily focus on learning representations of input modalities.
However, this leaves out a critical part of embodied AI, interacting with the environment.
Rather than pre-training representations, pre-training can also learn models of behavior that account for agent actions.
One line of work pre-trains models with supervised learning to predict actions from sensor inputs on large interaction datasets and then fine-tune this model to specific downstream tasks~\cite{baker2022video}.
Other work learns skills or reusable behaviors from offline datasets that can adapt to downstream tasks~\cite{pertsch2020accelerating,gupta2019relay}.
Future work may explore how scaling dataset size, model size, and compute can pre-train behavioral policies better suited for fine-tuning on downstream tasks.

\subsection{World models and inverse graphics}
As previously discussed, semantic and free-space maps have been hugely successful in enabling high performance and efficient learning across embodied-AI tasks (\eg in navigation~\cite{chaplot2020learning} and rearrangement~\cite{trabucco2022mass}). These mapping approaches are successful as they provide a simple, highly-structured, model of the agent's environment that enables explicit planning. The simplicity of existing mapping approaches is also one of their major limitations: as embodied tasks become more complex they require agents to reason about new semantic categories and new types of interaction (\eg arm-based manipulation).
Extending existing approaches to include new capabilities is generally possible but non-trivial, often requiring substantive human effort. For instance, a 2D free-space mapping approach successful for PointGoal Navigation~\cite{chaplot2020learning} was explicitly extended to include semantic mapping channels so as to enable training agents for ObjectGoal Navigation~\cite{chaplot2020object}. 
These challenges in mapping raise an important question: how can we build flexible models of an agent's environment that can be used for general purpose task planning? We identify two exciting directions toward answering this question: end-to-end trainable world models and game-engine simulation via inverse-graphics.

At a high-level, a world models $W$ is a function that, given the state of the environment $s_t$ at time $t$ and an agent action $a$, produces a prediction $W(s,a)=\widehat{s}_{t+1}$ of the state of the world at time $t+1$ if the agent were to take action $a$~\cite{ha2018worldmodels}. Iterative applications of the world model can thus be used to simulate agent trajectories and, thus, for model-based planning. As may be expected, building and training world models made challenging by several factors: (1) generally full state information ($s_t$) is not available as agent's have access only to partial, egocentric, observations, (2) the dynamics of an environment are frequently stochastic and thus cannot be predicted deterministically, (3) many details encoded in a state are irrelevant to task completion (\eg minor color or texture variations of objects) and attempting to predict these details needlessly complicates training, and (4) collecting high-quality training data for the end-to-end training of world models may require the design of increasingly complex physical states (\eg a tower of plates to be knocked over). While more work is needed before world models will become a ubiquitous tool for embodied AI agents, recent work has shown that world models can be successfully used to training agents to play Atari games~\cite{hafner2021discreteworldmodels} and to build navigation-only models of embodied environments~\cite{koh2021pathdreamer}.

As world models are meant to be broadly applicable and learned from data, they frequently eschew inductive biases and use general purpose architectures. The disadvantage of this approach is clear: we have well-understood models of physics that should not have to be re-learned from data for every task. Moreover, we have simulators designed explicitly to simulate 3D objects and their physical interactions, video game engines. These observations suggest another approach: rather than learning an implicit world model, can we use techniques from inverse-graphics to back-project an agent's observations to 3D assets within a scene in a game engine? Once this back-projection is complete, the game engine can be used to perform physical simulations and planning. This approach, which can be thought of as world modeling with strong inductive biases, has used successfully to build models of intuitive physics in constrained settings~\cite{wu2017learningtoseephysics}. While this approach appears very promising it does present some challenges: (1) the problem of inverse graphics is especially challenging in this setting as de-rendered objects must be in physically plausible relationships with one another for simulation to be meaningful and (2) game-engines are, generally, non-differentiable and can be slow. Nevertheless, this approach of explicitly bringing our understanding of physical laws to world models seems a promising direction toward building embodied models that can physically reason and plan.

\subsection{Simulation and Dataset Advances}

One factor towards improving the reliability and scope of embodied AI research in the future will be the continued improvement of simulation capabilities and realism, and increase in the scale and quality of 3D assets used in simulation.
Repeatable, quantitative analysis of embodied AI systems at scale has been made possible through the use of simulation.
As research in embodied AI continues to grow and tackle increasingly complex problems within increasingly complex scenes, the needs placed on simulation environments and assets  will increase. 

One important area of improvement for simulation environments is physics realism during agent-object interaction. Past simulation environments have solidly supported both abstracted~\cite{ai2thor,puig2018virtualhome} and rigid-body physics-based agent interactions~\cite{shen2020igibson, szot2021habitat, gan2020threedworld, ehsani2021manipulathor}. There has been quite some progress in physics simulation of flexible material (rope, cloth, soft body)~\cite{lin2020softgym, seita_bags_2021}, fluids~\cite{fu2022rfuniverse}, and contact-rich interaction (e.g. nut-and-bot)~\cite{narang2022factory}, leveraging state-of-the-art physics engines like PyBullet~\cite{coumans2021} and NVIDIA's PhysX/FleX. Some environment like iGibson 2.0~\cite{li2021igibson} even attempts to go beyond kinodynamic simulation and use approximate models to simulate more complex physical processes such as thermodynamics. However, all of these simulations are still far from perfect and oftentimes face a grim trade-off between fidelity and efficiency. More efficient and realistic simulation of physical interaction of agents with all elements of their environment can greatly assist in the applicability of embodied AI trained using simulation, to solving real-world problems.

With the prevalence of vision sensors for solving problems, the need for increased visual realism has also become imperative for research that is to translate to the real world. This has been aided in recent years through aspects like new graphics technology like real-time ray tracing.
An example of how these advances can improve visual realism can be found within iterations of the RVSU challenge~\cite{hall2020robotic}  %
that recently migrated to NVIDIA's Isaac Omniverse\footnote{see  \url{https://developer.nvidia.com/blog/making-robotics-easier-with-benchbot-and-isaac-sim/} for details}. Yet, the rendering speed can still become a bottleneck as the number of objects and light sources increase in the scenes. 

Aside from advances in computer graphics, visual realism also relies on high-quality 3D assets of scenes and objects. It has been a standard practice for embodied AI researchers to benchmark navigation agents in large-scale static scene datasets like Matterport3D~\cite{matterport3d}, Gibson~\cite{xiazamirhe2018gibsonenv}, and HM3D~\cite{ramakrishnan2021habitat}. On the other hand, interactive scenes have been quite limited. iGibson 2.0~\cite{li2021igibson} provides fifteen fully interactive scenes with added clutter that aim to capture the messiness of the real world, and Habitat 2.0~\cite{szot2021habitat} also similarly converts a subset of an existing static dataset~\cite{replica19arxiv} to become fully interactive. ProcTHOR~\cite{deitke2022procthor} recently attempted to scale up the effort and procedurally generate fully interactive scenes with realistic room structures and object layout.

Many object datasets have been proposed and heavily utilized by embodied AI researchers in the past years~\cite{chang2015shapenet,mo2019partnet,xiang2020sapien,calli2017yale,srivastava2022behavior,downs2022google,collins2022abo}. Although increased scale and quality has been the general trend for these datasets, it still remains extremely costly to make them useable for interactive tasks. For example, most of the objects in these datasets do not support interaction, such as the ability to open cabinets. Such work not only requires modifying meshes, but also requires a tremendous amount of annotation to provide part-level and articulation annotation, as was done in the PartNet and PartNet-Mobility datasets~\cite{mo2019partnet, xiang2020sapien}. Similarly, it requires additional annotation and mesh editing to support object states (\eg whether the object is cookable, sliceable) for the BEHAVIOR dataset~\cite{srivastava2022behavior} or in AI2-THOR~\cite{ai2thor}. Yet, these annotations are essential as we ramp up the complexity of embodied AI tasks.

Another important aspect of realistic simulation is its multimodal nature, one of the most important ones is auditory perception. Existing acoustic simulation like SoundSpaces~\cite{chen_soundspaces_2020} allows the agent to move around in the environment with both visual and auditory sensing to search for a sounding object. However, it pre-computes the room impulse response (RIR) based on scene geometry and can't be configured. Recent work like SoundSpaces 2.0~\cite{chen22soundspaces2} (Fig.~\ref{fig:ss2}) extended the simulation to make it continuous, configurable and generalizable to arbitrary scene datasets, which enables the agent to explore the acoustics of the space even further.

In addition, tactile sensing is also super important to future simulation environments. As these sensors become more cost-efficient, robots will likely be equipped with these new sensing capabilities in the foreseeable future. Researchers have made tremendous progress in tactile simulation~\cite{narang2021sim, agarwal2021simulation} in the past years, which can unlock tremendous potential for multi-modal embodied AI research.

\subsection{Sim2Real Approaches}

As the embodied AI community grows, and benchmarks in simulation continue to improve, a fundamental question that remains is: how well does this progress translate to the real world? Towards answering this question, the embodied AI community has made significant efforts in 1) building infrastructure to facilitate sim2real transfer on hardware, 2) providing support for researchers across the world to evaluate policies in the real-world, and 3) developing sim2real adaptation techniques. 

Significant advances have been made in recent years on real-world hardware targets, with the emergence of low-cost robots for evaluation \cite{pyrobot2019,kemp2022design} and open-source infrastructure for sim2robot deployment \cite{habitat2020sim2real, talbot2020benchbot, deitke2020robothor}. These advances have lowered the barrier to entry for robotics, and enable the embodied AI community to evaluate the performance of various research algorithms both in simulation and on real-world robots. Currently, each approach is limited to a specific simulator or a limited set of robot platforms. A key future direction is for these translation technologies to become ubiquitous interfaces, with support for any simulator or physical robot platform required by the researcher.

By comparing the performance of policies in simulation and the real-world, researchers are able to identify flaws in the simulator design that lead to poor sim2real transfer \cite{habitat2020sim2real}, and develop novel methods to overcome the sim2real gap. Common approaches for bridging the sim2real gap include domain randomization \cite{tobin2017domain, anderson2020sim}, or domain adaptation, a technique in which data from a source domain is adapted to more closely resemble data from a target domain. Prior works leveraged GAN techniques to adapt the visual appearance of objects from sim-to-real \cite{rao2020rl}, and other works built models \cite{truong2021bi, truong2022kin2dyn, deitke2020robothor}, or learned latent embeddings of the robot’s dynamics \cite{truong2020learning, kumar2021rma, yu2017preparing} to adapt to the actuation noise found in the real world. Models of real-world camera and actuation noises have since been integrated into simulators, and included as part of the Habitat, RoboTHOR, RVSU and iGibson Challenges, thereby improving the realism of the challenge and decreasing the sim2real gap. Continuing this close integration between real-world evaluation and improving simulators and benchmarks will help accelerate the speed of progress in robotics research.

A final future direction, is in addressing the differences between simulated and real-world sensorimotor interfaces. It is common currently for actuation to be broken into discretised chunks, and simulated sensor inputs treated the same as real-world inputs. While simulators and datasets will continue to advance, there will likely always be a difference between emulated and real-world sensorimotor experiences. Research approaches that leverage simulated data to learn policies, then embrace the limitations of these policies when transferring to real-world scenarios, have begun to emerge in recent years \cite{rana2021zero}. This is a start, but approaches like these will need to be expanded upon in the future.

\subsection{Procedural Generation}
In embodied AI, procedural generation can be used to create environments with respect to some priors. The purpose is often to scale up the diversity of data available used during training. The use of large-scale training data has resulted in models that are increasingly more powerful and generalizable across AI~\cite{ramesh2021zero,alayrac2022flamingo,chowdhery2022palm,radford2021learning}. Yet, most works in embodied AI often suffer from massive overfitting to the training scene datasets. For instance, in the RoboTHOR and Habitat ObjectNav challenges alone, it is not uncommon to obtain near 100\% success on the 100 or so scenes seen during training while only obtaining 30-50\% success when evaluating on unseen scenes.

Early work in embodied AI either trained agents on hand-designed scenes created by 3D artists~\cite{zhu2017target} or from static 3D scans of real-world environments~\cite{chang2017matterport3d}. However, there are several drawbacks for both approaches. Manually creating 3D scenes is incredibly time intensive work and requires graphics experts to create 3D assets, possibly make them interactive, and arrange the objects to construct scenes. It took 3D artists about 32 hours to develop each house of the ArchitecTHOR dataset~\cite{deitke2022procthor}, and results in present-day simulators, with scenes designed by 3D artists, to only having on the order of 100 scenes available for training \cite{deitke2020robothor, li2021igibson, szot2021habitat}. Static 3D scans do not support interacting or manipulating objects and may be incredibly time consuming to annotate semantically, capture the scenes, and clean up the meshes. To create HM3D-Semantics~\cite{ramakrishnan2021habitat}, it took over 100 hours to semantically annotate each of 120 scenes to make them useable for ObjectNav. Thus, it is incredibly difficult to scale the creation of both hand-designed scenes and 3D scanned scenes.

\begin{figure}[ht!]
    \centering
    \includegraphics[width=0.47\textwidth]{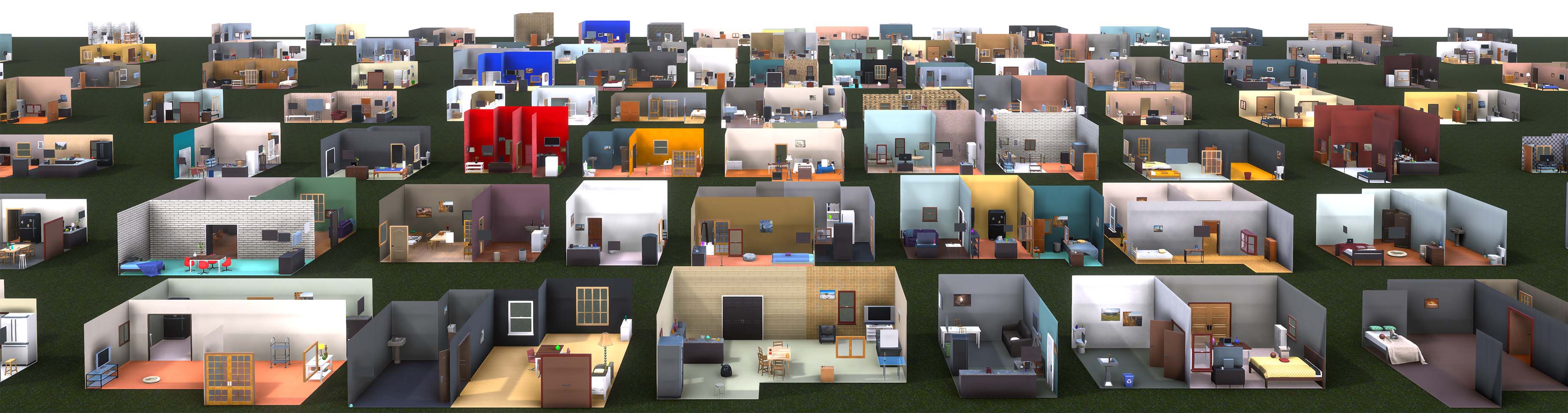}
    \caption{Examples of Procedurally generated houses from ProcTHOR~\cite{deitke2022procthor}.}
\end{figure}

Procedurally generating environments offers an alternative approach towards scaling data in embodied AI, which randomly samples scenes with respect to a prior distribution. ProcTHOR~\cite{deitke2022procthor} procedurally generates 10K houses to train embodied agents, and achieves remarkable generalization results on many downstream navigation and interaction tasks~\cite{ramakrishnan2021habitat, deitke2020robothor, ehsani2021manipulathor, weihs2021rearrangement}. At a high-level, each house is generated by sampling a floorplan (defining which rooms appear in a house and where) and then sampling the placement of objects in rooms of the house. The objects largely come from AI2-THOR's asset database, making them interactive. All the objects are modularly placed in each scene, which also provides semantic annotations of the scenes for free. Procedural generation also allows for sampling scenes with respect to preferences. For example, if I want to train an ObjectNav agent to find a basketball in a home or to operate in a room with many mirrors, it is significantly easier to procedurally generate environments that fit such criteria than it is to either build such environments from scratch or find and scan such environments in the real world. \cite{team2021open,baker2019emergent,fu2021minimizing} have also shown impressive generalization results from procedurally generating more simplistic embodied environments. We suspect the trend of procedurally generating environments to continue to grow in the years to come.

\subsection{Generalist Agents}

Embodied agents that can learn from many types of inputs and produce many different types of outputs offer a promising approach to generalize to interacting with humans and being able to quickly adapt to new tasks. Here, we may want our agents to be able to learn from watching videos, viewing a geographic map, reading a tutorial, or listening to somebody talk, and be able to communicate through navigation or manipulation actions, text, or voice. The promise of generalist agents in embodied AI is that they should benefit from knowledge transfer between tasks and modalities, while being much easier to instruct and adapt to new tasks.

An emerging phenomenon in generative NLP models, such as GPT-3~\cite{brown2020language} and PALM~\cite{chowdhery2022palm}, is that they can be used to solve arbitrary NLP tasks in a 0-shot setting by prompting the models with language tokens as input, and having it generate language tokens from the same vocabulary as output. Such prompting can be used to evaluate the models on many tasks, such as question answering, summarization, and mathematical reasoning. However, in the realm of computer vision, the input and output modalities are incredibly different. For example, optical flow models might input a video and output a flow mask of each frame; object detection models input an image and output a set of bounding boxes with their associated classes; and image generation models input a text description, and output an RGB image. Thus it is much harder to build unified computer vision models. Building unified models to achieve arbitrary tasks in embodied AI is similarly difficult due to the many input and output modalities that are possible.

\begin{figure}[ht!]
    \centering
    \includegraphics[width=0.47\textwidth]{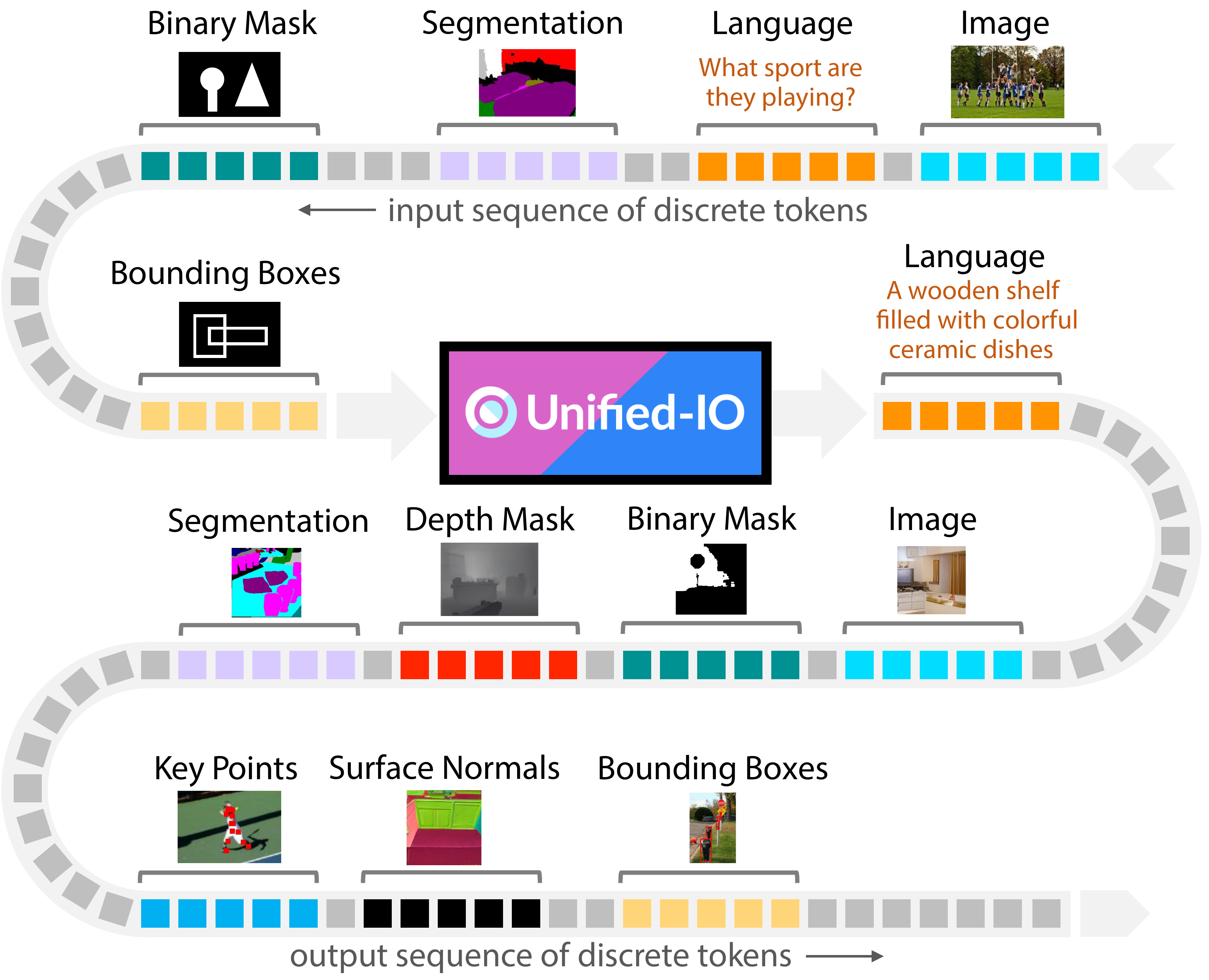}
    \caption{Unified-IO~\cite{lu2022unified} is a generalist model in computer vision, which can be used to solve tasks with many different input and output modalities.}
\end{figure}

Gato~\citep{reed2022generalist} is the first big attempt at building a unified model that works for embodied agents. It consists of a single transformer agent that was trained on a wide variety of vision, language, control, and multi-modal tasks. While currently the results indicate that Gato does not fully benefit from a shared framework, in the future, with sufficient scaling and better problem formulation, the authors hypothesize that it will achieve significant gains in generalization. In computer vision, there has recently been a line of work exploring unified models, including Unified-IO~\cite{lu2022unified}, Perceiver-IO~\cite{jaegle2021perceiver}, and UViM~\cite{kolesnikov2022uvim}. These models are able to perform a wide variety of tasks in both vision and language out of the box, including performing well with image impainting, segmentation, and visual question answering. Moreover, their results begin to show models that transfer knowledge between tasks. In the years to come, we suspect the rise of unified models will continue advancing, growing to support 0-shot task completion for many more tasks in embodied AI.

In parallel to multi-task learning, recent works have also experimented with re-purposing existing pre-trained models. In Socratic Models~\citep{zeng2022socratic}, GPT-3 (a language model)~\citep{brown2020language}, CLIP (a vision-language model)~\citep{radford2021learning}, and CLIPort (a vision-language-action model)~\citep{shridhar2022cliport} are used together to solve tabletop pick-and-place tasks, with language as the common interface to prompt the pre-trained models.
Similarly, in SayCan~\cite{ahn2022can}, a language model is used to generate high-level plans that be can executed with pre-trained action skills. Likewise, in Inner Monologue~\citep{huang2022inner} and ALFWorld~\cite{ALFWorld20}, textual state descriptions as used as a medium for sequential decision-making.

Overall, while \textit{generalist} agents are less prevalent in Embodied AI, in the near future, we might see greater consolidation of tasks and agent architectures following similar progress in vision and NLP.

\subsection{Multi-Agent \& Human Interaction}

Analogous to the social learning in humans, it is desirable that embodied agents can observe, learn from, and collaborate with other agents (including humans) in their environment. The advanced and realistic simulated environments being developed for Embodied AI research will serve as virtual worlds for agent-agent and human-agent interaction. The two pillars for social, multi-agent, and human-in-the-loop embodied agents are (1) accurately simulating a subset of agent and human behavior relevant to a given embodied task and (2) creating realistic benchmarks for multi-agent and human-AI collaboration.

\begin{figure}[ht!]
    \centering
    \includegraphics[width=0.47\textwidth]{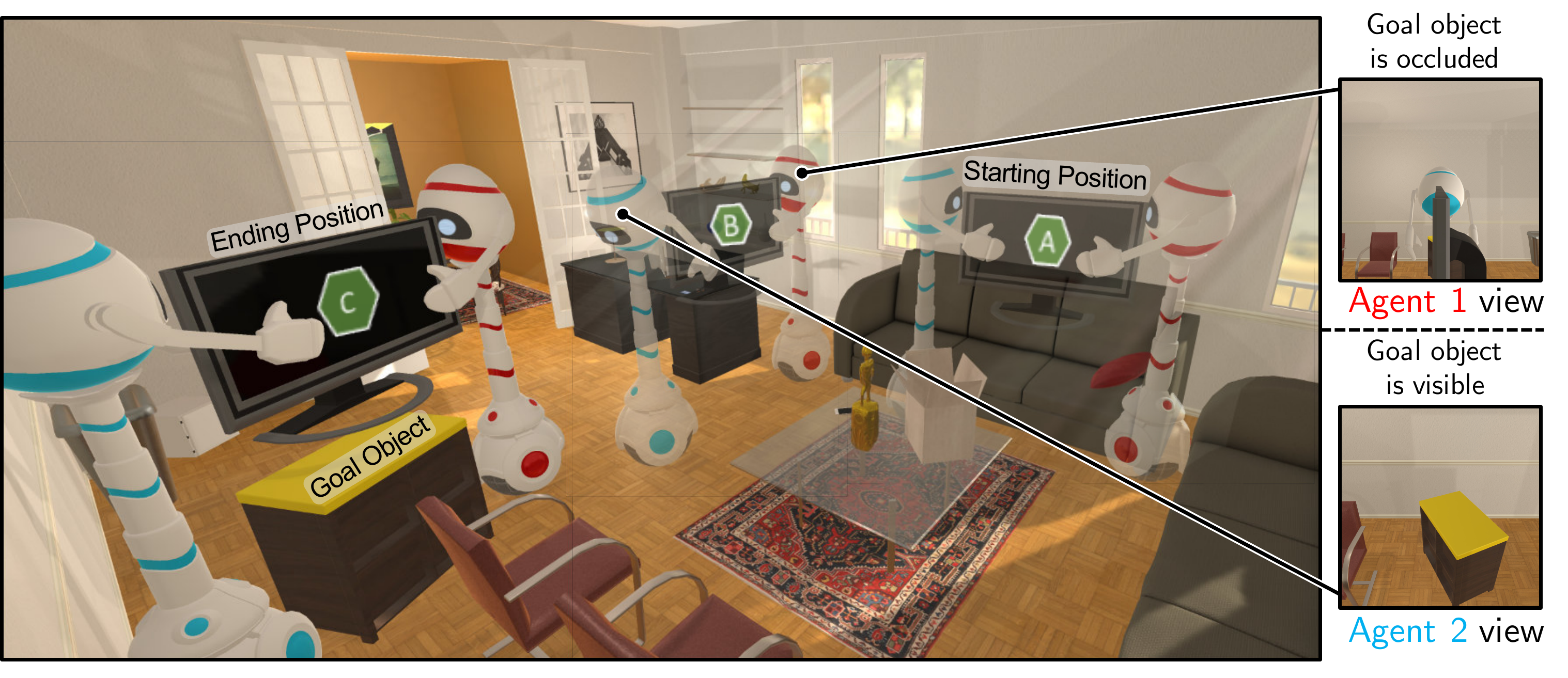}
    \caption{Furniture Moving~\cite{jain2020cordial} is a collaborative multi-agent task for agents to move a heavy furniture item.}
\end{figure}

Immersing humans in simulation creates an opportunity for a new class of experiences and user studies that involve human-virtual agent interaction, data collection of human demonstrations at scale in controlled environments, and creation of events and visualizations that are impossible or irreproducible in real scenarios. Some examples towards this goal include VirtualHome~\cite{puig_cvpr18} where programs are collected and created to model human behaviors along with animated atomic actions such as walk/run, grab, switch-on/off, open/close, place, look-at, sit/standup, touch. 
TEACh~\cite{teach} collects both human instructions, demonstrations, and question answers from human who interact with the simulator through a web interface~\cite{teach}, while BEHAVIOR uses virtual reality to collect high-fidelity human demonstrations directly in the action space of a simulated robot agent~\cite{srivastava2022behavior}.
To train policies, modeling the task-relevant aspects of human behavior is of prime focus. 
In challenges such as SocialNav, human agents are simulated following a simple interaction model that considers interactions between agents. 
Looking forward, with robust motion solutions models~\cite{rong2021frankmocap,lugaresi2019mediapipe} and human behavior animation~\cite{won2020scalable,2021-TOG-AMP}, emulating from large-scale human-activity datasets~\cite{gu2018ava,smaira2020short,damen2022rescaling,grauman2022ego4d} is an exciting prospect for modeling human behaviors in simulation. To  train and transfer these policies to the real world, we must develop low-shot approaches and realistic benchmarks to learn socially intelligent agents.

\begin{figure}[ht!]
    \centering
    \includegraphics[width=0.47\textwidth]{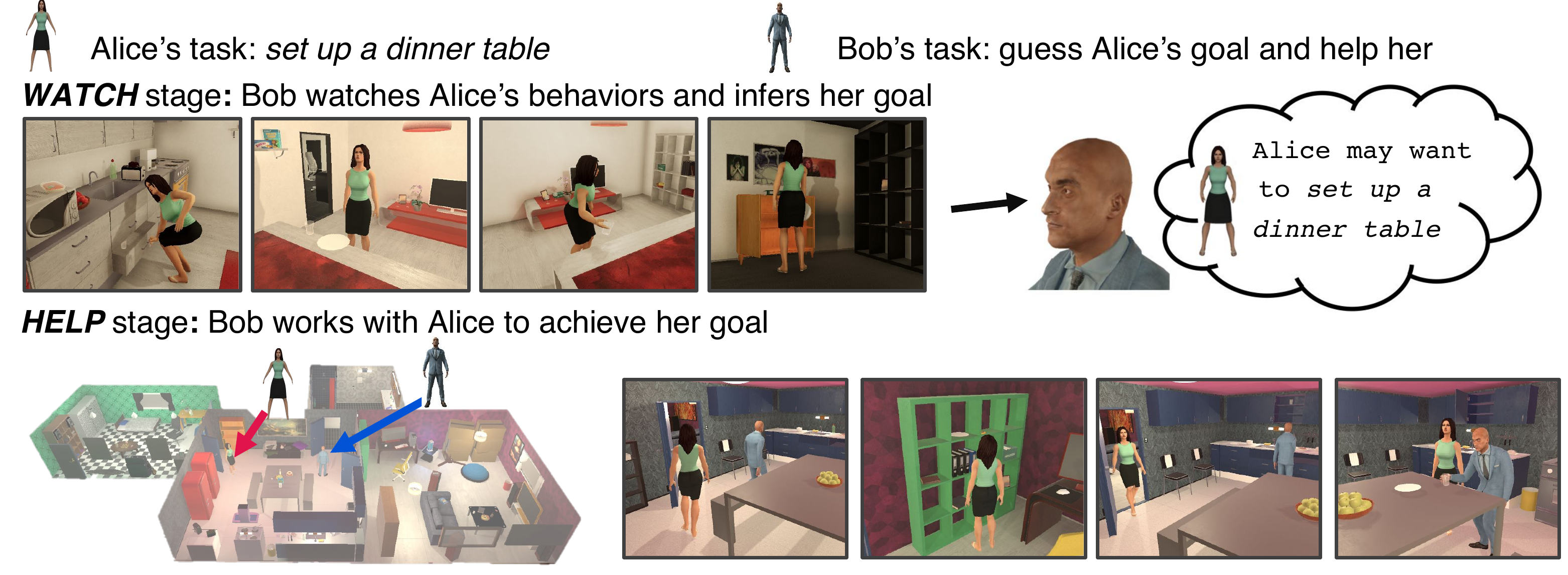}
    \caption{Watch and Help encourages social intelligence where an agent learns in the presence of human-like teachers (image credits: Puig~\etal~\cite{puig2021watchandhelp}).}
\end{figure}

Several benchmarks have helped make progress within the space of multi-agent and social learning in embodied AI. Within AI2-THOR, collaborative task completion~\cite{jain2019two} and furniture moving~\cite{jain2020cordial} were one of the first benchmarks for multi-agent learning in embodied AI, focussed on task that cannot be done by a single agent. While abstracted gridworlds~\cite{jain2021gridtopix} provide a faster training ground for such tasks, efficiently going beyond 2-3 agents models with high visual fidelity is challenging. Emergent communication~\cite{Patel_2021_ICCV} and emergent visual representations~\cite{weihs2020learning} show  examples of learning heterogeneous agents possessing specialized skills. SocialNav in iGibson presents early steps towards robot learning for mobility around humans and other moving objects within the environment. Within VirtualHome, the watch-and-help~\cite{puig2021watchandhelp} benchmark will enabled few-shot learning of policies that can interact with a human-like agent to replicate demonstrations in an unseen environment.

Overall, simulated environments offer a scalable platform for procedural training and testing of interactive policies, potentially addressing some of the limiting challenges inherent to research on human interaction: scaling up with safety and speed, standardize environments to support reproducible research, and procedural testing and benchmarking of a minimum set of tests before deploying on real robots. Progress on all these fronts requires the integration and convergence of contributions from diverse fields such as graphics, animation, and simulation, towards fully functional, realistic and interactive virtual environments. 

\subsection{Impact of Embodied AI}

Whether in simulation or reality, embodied AI research focuses on embodied tasks in the hope of delivering on the fundamental promise of AI: the creation of embodied agents, such as robots, which learn, through interaction and exploration, to creatively solve challenging tasks within their environments.
Many embodied AI researchers believe that creating intelligent agents that can solve embodied tasks will produce outsized real-world impacts.
Increasingly capable robotic platforms and effective sim-to-real techniques make it easier to transfer learned policies to the real world.
Even small advances at interesting embodied tasks could serve as the foundation for technologies that could improve the lives of people with disabilities or free able-bodied humans from mundane tasks.
However, these advances, as with all automation, could result in disruptions such as the elimination of jobs or disempowerment of individuals.
We must be careful to ensure that the benefits of embodied AI become available to all and do not reinforce inequality.
Therefore, the embodied AI community has promoted discussion of these issues in the hope that it will guide us towards more equitable solutions.

\section{Conclusion}

In this paper, we presented a retrospective on the state of Embodied AI research. We discussed 13 different challenges that make up a testbed for a suite of embodied navigation, interaction, and vision-and-language tasks. Over the past 3 years, we observed large-scale training, visual pre-training, modular and end-to-end training, and visual \& dynamic augmentation as common approaches to many of the top challenge entries. We discuss improvements to pre-training, world models and inverse graphics, simulation and dataset advances, sim2real, procedural generation, generalist agents, and multi-agent \& human interaction as promising future directions in the field.

\section*{Contributions}

\paragraph{Matt Deitke} led the planning, outline, and coordination of the paper; worked on the abstract, introduction, \& conclusion and worked on the ObjectNav section, the large-scale training section, the procedural generation, and the generalist agents section.

\paragraph{Yonatan Bisk} said we should do this, attended a few planning meetings, but then delegated and Matt really ran with it.

\paragraph{Tommaso Campari} co-wrote the section on Multi-ObjectNav challenge.

\paragraph{Devendra Singh Chaplot} worked on Habitat Challenge sections and the end-to-end vs modular subsection.

\paragraph{Changan Chen} worked on the audio-visual navigation section and the simulation and dataset advances section.

\paragraph{Claudia P\'{e}rez-D'Arpino} worked on the Introduction, Interactive and Social PointNav, and the Multi-Agent \& Human Interaction sections.

\paragraph{Anthony Francis} worked on the Introduction, What Is Embodied AI, and Sim to Real sections, and edited other sections.

\paragraph{Chuang Gan} worked on the rearrangement challenges section.

\paragraph{David Hall} worked on the RVSU challenge sections and provided some editing on the simulation and dataset advances section.

\paragraph{Winson Han} created the Figure 1 cover graphic.

\paragraph{Unnat Jain} worked on audio-visual navigation, multi-object navigation, and multi-agent sections.

\paragraph{Jacob Krantz} worked on the challenge section on Navigation Instruction Following.

\paragraph{Chengshu Li} worked on the Interactive and Social PointNav section and the Simulation and Dataset Advances section in Future Directions.

\paragraph{Sagnik Majumder} worked on the audio-visual navigation section.

\paragraph{Roberto Mart\'{i}n-Mart\'{i}n} worked on the What Is Embodied AI section, and the Interactive and Social PointNav section.

\paragraph{Sonia Raychaudhuri} co-wrote the section on Multi-ObjectNav challenge.

\paragraph{Mohit Shridhar} worked on the interactive instruction following section for challenge details and the generalist agents section for future directions. 

\paragraph{Niko S\"{u}nderhauf} worked on the RVSU challenge sections.

\paragraph{Andrew Szot} worked on the pre-training section for future directions.

\paragraph{Ben Talbot} worked on the RVSU challenge sections and Sim2Real Approaches advances section.

\paragraph{Jesse Thomason} worked on the interactive instruction following and interactive instruction following with dialog sections of the challenge details, and the multi-agent \& human interaction section of future directions.

\paragraph{Alexander Toshev} worked on Social and Interactive Navigation section.

\paragraph{Joanne Truong} worked on the PointNav section for challenge details, and the Sim2Real approaches section for future directions.

\paragraph{Luca Weihs} worked on the rearrangement section for challenge details, the visual pre-training section for common approaches, and the world models and inverse graphics section for future directions.

\paragraph{Dhruv Batra, Angel X. Chang, Kiana Ehsani, Ali Farhadi, Li Fei-Fei, Kristen Grauman, Aniruddha Kembhavi, Stefan Lee, Oleksandr Maksymets, Roozbeh Mottaghi, Mike Roberts, Manolis Savva, Silvio Savarese, Joshua B. Tenenbaum, Jiajun Wu} advised and provided feedback on the draft, workshop, and/or challenges.

\section*{References}

\singlespace
\renewcommand{\section}[2]{}
\bibliographystyle{plain}
\bibliography{strings,main}

\end{document}